\documentclass{article} 
\PassOptionsToPackage{sort}{natbib}
\usepackage[main,final]{neurips_2026}
\setcitestyle{numbers,square,citesep={,}}

\usepackage{amsmath,amsfonts,bm}

\def\eqref#1{equation~\ref{#1}}

\def\1{\bm{1}}

\DeclareMathAlphabet{\mathsfit}{\encodingdefault}{\sfdefault}{m}{sl}
\SetMathAlphabet{\mathsfit}{bold}{\encodingdefault}{\sfdefault}{bx}{n}

\usepackage{xspace}

\usepackage{array, boldline, makecell}
\usepackage[svgnames, table]{xcolor}
\usepackage{multirow}
\usepackage{booktabs} 
\usepackage[table]{xcolor}
\usepackage{enumitem}
\usepackage{algorithm}
\usepackage{algpseudocode}  
\usepackage{wrapfig}
\usepackage{placeins}

\usepackage[utf8]{inputenc} 
\usepackage[T1]{fontenc}    
\usepackage{hyperref}       
\usepackage{url}            
\usepackage{amsfonts}       
\usepackage{nicefrac}       
\usepackage{microtype}      
\usepackage{xcolor}         
\usepackage{amsmath}
\usepackage{amssymb}
\usepackage{mathtools}
\usepackage{amsthm}
\usepackage{graphicx}
\usepackage{subcaption}

\makeatletter
\NewDocumentCommand{\supptitle}{s}{
\twocolumn[{
  \begin{center}
    \vspace*{-0.3cm}
    \rule{\textwidth}{0.05cm}\\[0.1cm]
    \textbf{- Appendix -}\\[0.2cm]
    {\Large \textbf{\mytitle}}\\[0.1cm]
    \rule{\textwidth}{0.05cm}\\[0.3cm]
  \end{center}
}]
}
\makeatother

\definecolor{LightCyan}{rgb}{0.88,1,1}
\definecolor{Blue}{rgb}{0, 0.5, 1}
\definecolor{Green}{rgb}{0.0, 0.8, 0.0 }
\definecolor{Red}{rgb}{0.95, 0.55, 0.6}
\definecolor{Skyblue}{rgb}{0.6, 0.6, 0.95 }
\definecolor{Beige}{rgb}{0.96, 0.96, 0.86}

\newcommand{\system}{\texttt{SAGA}}

\usepackage[normalem]{ulem}
\useunder{\uline}{\ul}{}

\usepackage{tcolorbox}
\tcbuselibrary{skins, breakable}

\usepackage[hang,flushmargin]{footmisc}

\newcommand{\rev}[1]{{\color{black}#1}}
\newenvironment{revision}{\par\begingroup\color{black}}{\par\endgroup}

\usepackage{needspace}

\usepackage{hyperref}
\usepackage{url}
\usepackage{etoc}

\title{Stage-wise Attention-Guided Region Sequencing for Adversarial Attacks on Large Vision-Language Models}

\author{%
Jaehyun Kwak \\
KAIST
\And
Nam Cao \\
KAIST
\And
Boryeong Cho \\
KAIST
\And
Segyu Lee \\
KAIST
\AND
Sumyeong Ahn\textsuperscript{\textdagger} \\
KENTECH
\And
Se-Young Yun\textsuperscript{\textdagger} \\
KAIST
}

\hypersetup{hidelinks}

\begin{document}

\maketitle
\begingroup
\renewcommand{\thefootnote}{\fnsymbol{footnote}}
\footnotetext[2]{Corresponding authors.}
\endgroup
\etocdepthtag.toc{main}

\begin{abstract}
    Targeted adversarial attacks on Large Vision--Language Models (LVLMs) test whether small image perturbations can steer model responses toward attacker-specified content. Under the standard $L_\infty$ constraint, targeted attacks become a regional perturbation budget allocation problem: attack success depends not only on the perturbation objective, but also on which regions receive updates and in what order. Existing localized attacks improve over global perturbations but rely on stochastic spatial sampling, often updating weakly influential regions. \rev{We address this limitation through an attention-based analysis linking regional attention to adversarial sensitivity and examining how hotspot perturbations redistribute attention toward subsequent salient regions.} These findings motivate attention-guided region sequencing, which begins from dominant hotspots and progressively moves the update support toward next-salient regions. Based on these principles, we propose Stage-wise Attention-Guided Attack (\system{}), a black-box region-sequencing framework that uses a fixed attention map from an open-source LVLM to guide perturbation updates without accessing target-model parameters, gradients, or attention maps. \rev{Across fourteen closed-source and open-source LVLMs, \system{} achieves state-of-the-art attack success rates and the best overall imperceptibility, with relative ASR gains of 40.0\% on Gemini 3 Pro and 30.2\% on DeepSeek V4.1 Flash over the strongest baseline for each target. It also reduces LPIPS by 13.8\% relative to the best baseline on this metric.}
    \rev{The source code is available at \url{https://github.com/jaehyun-kwak/SAGA}.}
\end{abstract}

\begin{figure}[h]
    \centering
    \includegraphics[width=1.0\columnwidth]{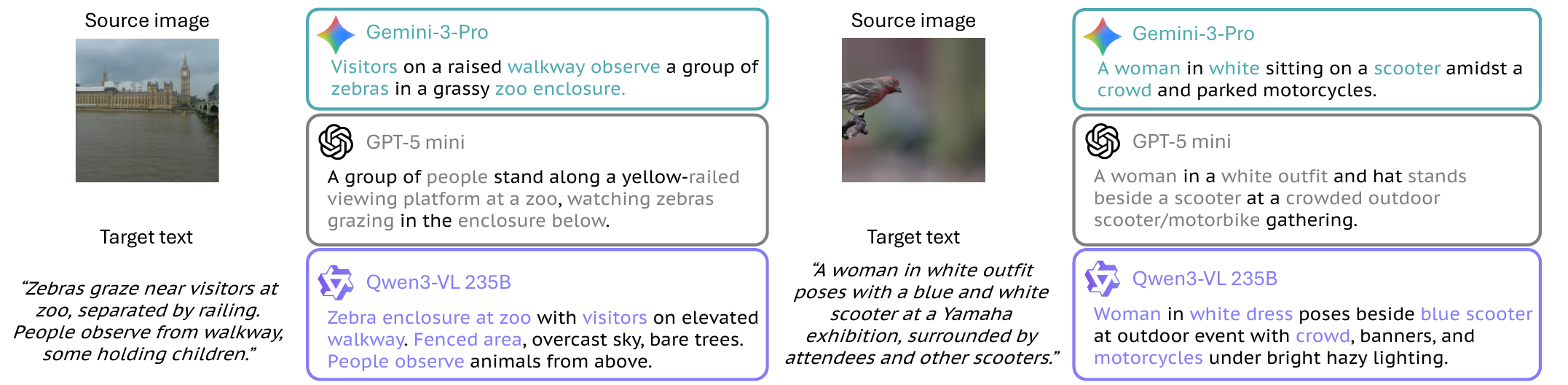}
    \caption{\rev{Example captioning responses from commercial and open-source LVLMs to images attacked by \system{}. Image perturbations can redirect generated descriptions toward the target text.}}
    \label{fig:intro_attack_example}
    \vspace{-1.7em}
\end{figure}

\section{Introduction}
\label{sec:intro}

Large Vision--Language Models (LVLMs) have shown strong capabilities in image captioning~\citep{li2022blip}, visual question answering~\citep{kuang2025natural}, and complex multimodal reasoning~\citep{chen2024spatialvlm}.
However, their growing use in multimodal systems raises important robustness and safety concerns~\citep{pathade_invisible_2025, clusmann_prompt_2025, xu2024shadowcast, lyu2024trojvlm, niu_jailbreaking_2024, ICLR2024_83170cce, zhao2023evaluating, dong_how_2023}.
We focus on targeted adversarial attacks on LVLMs, where small perturbations to a source image steer model responses toward attacker-specified content~\citep{zhao2023evaluating, dong_how_2023}.
Figure~\ref{fig:intro_attack_example} illustrates this threat across both commercial and open-source LVLMs.

Imperceptibility is central to this threat model. 
The standard $L_\infty$ constraint limits the maximum change allowed at each pixel, making visible artifacts less likely~\citep{goodfellow2014explaining, madry2017towards}.
\rev{Under this constraint, each pixel has a finite update range. Allocating updates to weakly influential regions can increase distortion without effectively steering the model toward the target text.}
Thus, an effective imperceptible attack must decide not only how to optimize perturbations, but also how to allocate the finite update budget across an ordered sequence of spatial regions.

\rev{Localized perturbations can outperform global image manipulation~\citep{li2025a, jia2025adversarial}, suggesting that image regions differ in adversarial sensitivity. However, existing localized attacks sample regions stochastically across the image without explicitly identifying vulnerable regions or ordering their updates. This can allocate the limited $L_\infty$ budget to regions that weakly influence visual grounding.}

\begin{figure*}[!t]
    \centering
    \begin{minipage}[t]{0.32\textwidth}
        \centering
        \includegraphics[width=\linewidth]{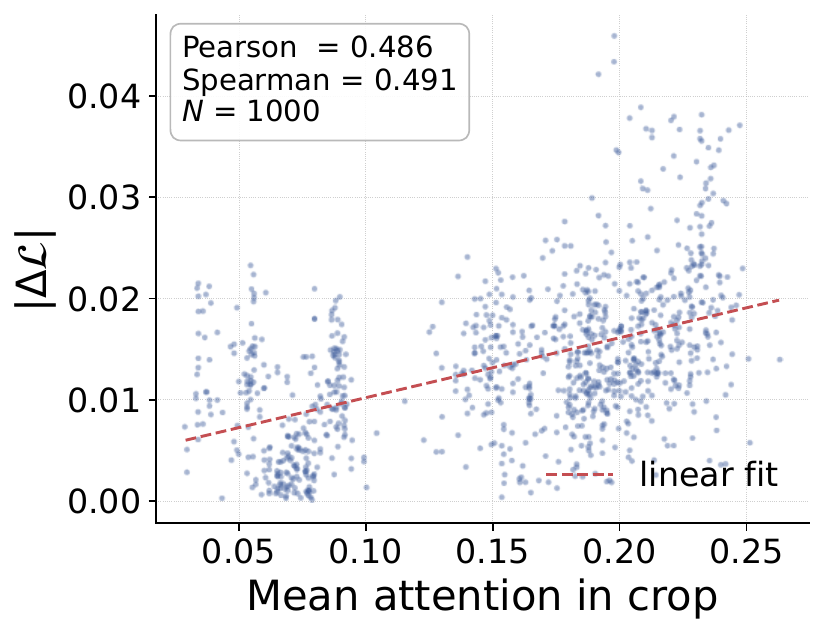}
        \caption{Pre-attack cross-modal attention vs. adversarial loss sensitivity on Qwen3-VL-8B. The strong positive correlation indicates that high-attention regions are more sensitive.}
        \label{fig:pre_corr_loss_attention}
    \end{minipage}
    \hspace{0.02\textwidth}
    \begin{minipage}[t]{0.64\textwidth}
        \centering
        \includegraphics[width=\linewidth]{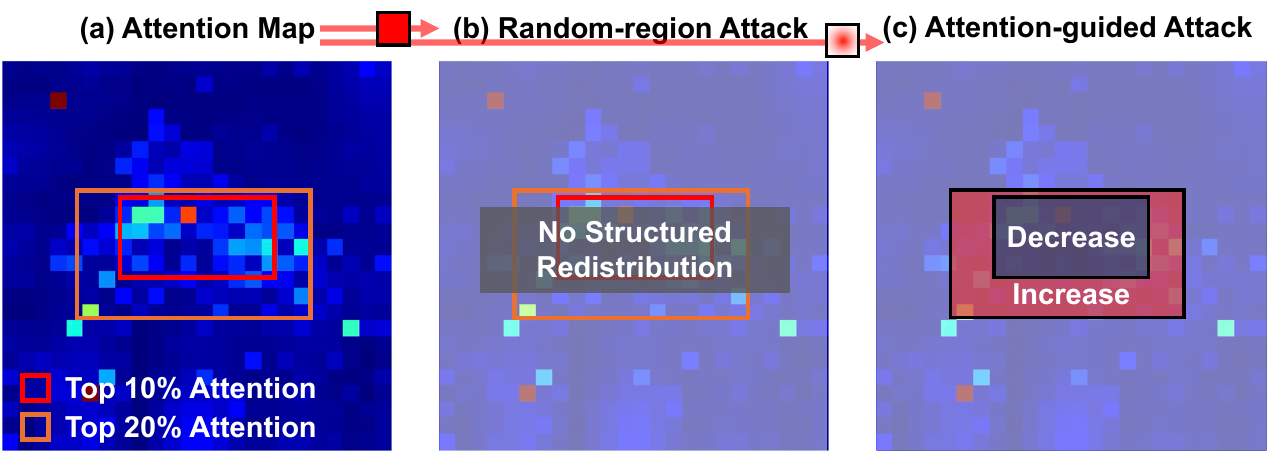}
        \caption{Attention redistribution under different attack strategies. (a) \textcolor{red}{Red} and \textcolor{orange}{orange} boxes denote the top 10\% and top 20\% attention regions. (b) Random-region attacks show no structured redistribution. (c) Attention-guided attacks decrease attention in the current hotspot and increase attention in the next-salient region.}
        \label{fig:pre_attention_shift}
    \end{minipage}
    \vspace{-1.5em}
\end{figure*}

We address this problem by analyzing localized attacks using an attention-based approach. 
\rev{First, our sensitivity analysis shows a positive correlation between regional attention scores and adversarial loss changes, with larger changes observed when perturbing high-attention regions (Section~\ref{subsec:motiv_where}, Figure~\ref{fig:pre_corr_loss_attention}).}
\rev{Second, we analyze attention redistribution under a local-effect approximation and observe larger mean attention gains in next-salient regions than in coldspots after hotspot attacks (Section~\ref{subsec:motiv_when}, Figure~\ref{fig:pre_attention_shift}).}
\rev{Together, these analyses motivate starting from high-attention regions and using pre-attack attention to order subsequent updates.}

Based on these principles, we propose Stage-wise Attention-Guided Attack (\system{}), a region-sequencing framework for perturbation-budget-efficient attacks on LVLMs. 
\system{} extracts a cross-modal attention map once from an open-source LVLM before optimization and constructs a sequence of high-attention hotspots. 
During optimization, it localizes perturbation updates to the ordered hotspot sequence while respecting the global $L_\infty$ constraint. 
\rev{\system{} requires no access to target-model parameters, gradients, or attention maps, allowing it to attack closed-source LVLMs in a black-box setting.}

We evaluate \system{} on \rev{fourteen} target LVLMs \rev{in the main comparison}, comprising \rev{seven} closed-source and \rev{seven} open-source models.
\rev{Across these models, \system{} consistently improves attack success rates over strong prior methods, with relative ASR improvements of 40.0\% on Gemini 3 Pro and 30.2\% on DeepSeek V4.1 Flash over the strongest baseline for each target. It also achieves the best results across all six imperceptibility metrics, including a 13.8\% reduction in LPIPS relative to the best baseline on this metric. Reusing a single pre-attack attention map achieves similar ASR to dynamic recomputation at substantially lower computational cost.}

Our contributions are summarized as follows:
\begin{itemize}[leftmargin=1.5em, itemsep=0pt, topsep=2pt]
    \item We formulate localized adversarial attacks on LVLMs as a perturbation budget allocation problem under an $L_\infty$ constraint, emphasizing region selection and region sequencing.
    \item We provide an attention-based characterization of localized attack dynamics, showing that cross-modal attention identifies vulnerable regions and that hotspot perturbations provide an ordering signal for subsequent updates.
    \item We propose \system{}, a stage-wise attention-guided region-sequencing attack that uses an attention map from a single open-source LVLM and transfers effectively to diverse closed-source and open-source target LVLMs.
    \item \rev{We evaluate \system{} on fourteen LVLMs, demonstrating state-of-the-art attack success, improved imperceptibility, and efficient perturbation budget use.}
\end{itemize}
We provide additional related work in Appendix~\ref{appendix:related_work} and discuss limitations and future directions in Appendix~\ref{appendix:limitations}.

\section{Preliminary}
\label{sec:preliminary}

\paragraph{Perturbation budget under $L_\infty$ constraint.}Adversarial attacks are formulated under an $L_\infty$ constraint to ensure that the resulting perturbations remain visually imperceptible. Under this setting, the objective is to generate an adversarial example $x_{\mathrm{adv}}$ that remains visually close to the original image $x_{\mathrm{orig}}$ while inducing the model to generate responses aligned with a target text description $y_{\mathrm{targ}}$.
\begin{equation}
\max_{x_{\mathrm{adv}}} \;
\mathcal{L}\bigl(f_{\mathrm{img}}(x_{\mathrm{adv}}), f_{\mathrm{txt}}(y_{\mathrm{targ}})\bigr),
\ \text{s.t.} \ \|x_{\mathrm{adv}} - x_{\mathrm{orig}}\|_\infty \le \epsilon .
\end{equation}
Here, $f_{\mathrm{img}}(\cdot)$ and $f_{\mathrm{txt}}(\cdot)$ denote the image and text encoders, respectively, and $\mathcal{L}$ denotes a cosine similarity loss.

The $L_\infty$ constraint enforces an independent $\epsilon$ bound on each pixel, yielding a finite per-pixel perturbation budget. This constraint introduces an implicit ordered allocation problem: allocating updates to weakly influential regions, or repeatedly updating a region whose influence has already diminished, can reduce the effectiveness of later perturbations.
As a result, the success of an $L_\infty$-bounded attack depends critically on which regions are updated and how the update support is sequenced across regions.

\section{Motivation: Region Selection and Sequencing}
\label{sec:motivation}

The $L_\infty$ constraint limits the per-pixel perturbation budget, making localized attack performance depend on how update regions are selected and ordered. 
We study this problem through cross-modal attention. 
Section~\ref{subsec:motiv_where} shows that high-attention regions provide effective attack targets. 
Section~\ref{subsec:motiv_when} then analyzes when attention redistribution provides a reliable signal for selecting the next update region.

\subsection{Region Selection: Attention Identifies Vulnerable Regions}
\label{subsec:motiv_where}

We first examine whether cross-modal attention can identify regions that are more sensitive to adversarial perturbations. 
For a given source image--target text pair $(x,y)$, we extract cross-modal attention maps from an open-source LVLM, such as Qwen3-VL~\citep{bai2025qwen3vltechnicalreport} or LLaVA~\citep{NEURIPS2023_6dcf277e}.
Our correlation analysis follows a three-step procedure:
\begin{itemize}[leftmargin=1.5em, itemsep=0pt, topsep=2pt]
    \item \textbf{Region sampling:} 
    We sample a spatial region $r \subset x$ from the source image.

    \item \textbf{Localized perturbation:} 
    We apply a single gradient step restricted to $r$ and measure the resulting adversarial loss change $\Delta \mathcal{L}_r$.

    \item \textbf{Attention--sensitivity mapping:} 
    We compute the pre-attack regional attention score $\overline{M}_r$ by averaging cross-modal attention values over visual patches inside $r$, and correlate $\overline{M}_r$ with $\Delta \mathcal{L}_r$ across sampled regions.
\end{itemize}

As shown in Figure~\ref{fig:pre_corr_loss_attention}, regional attention is positively correlated with adversarial loss sensitivity on Qwen3-VL-8B. 
We observe consistent positive correlations across other LVLMs and layers; details are provided in Appendix~\ref{appendix:correlation_exp}. 
These results suggest that cross-modal attention provides a useful cue for selecting vulnerable regions. 
Rather than relying on uniform or random region sampling, an attack can use high-attention regions as stronger targets under the same perturbation budget.

However, if the attack keeps updating only the same high-attention region, other regions can continue to support the original image semantics. 
This motivates the next question: how should the attack choose subsequent regions?

\subsection{\rev{Region Sequencing: Attention Redistribution under Hotspot Perturbations}}
\label{subsec:motiv_when}

Section~\ref{subsec:motiv_where} shows that high-attention regions are adversarially sensitive. 
We further observe that both hotspot-localized and random-region attacks can reduce the attention mass of the initial hotspot (see Appendix~\ref{app:negative_control} and Figure~\ref{fig:app_hotspot_suppression}). 
Given this decrease in hotspot attention, the central question is where the released attention moves and whether this redistribution is predictable.
\rev{We therefore analyze attention reallocation through softmax normalization under a local-effect approximation.}

Let $a_i=\mathrm{softmax}(z)_i$ denote the normalized attention weight of visual token $i \in \mathcal{V}$. 
For a region $S \subset \mathcal{V}$, let $m_S=\sum_{i\in S}a_i$ denote its total attention mass and $\bar a_S=m_S/|S|$ denote its mean attention score.
After perturbation, we use $a'_i$, $m'_S$, and $\bar a'_S$ for the corresponding post-attack quantities.
Let $\Delta_{\mathcal{H}} = m_{\mathcal{H}} - m'_{\mathcal{H}} > 0$ denote the attention mass reduced from the hotspot $\mathcal{H}$.

This decrease alone does not determine where the released attention moves. 
If the perturbation also changes non-hotspot logits, the attention gain of a non-hotspot region depends on the changes in its non-hotspot logits, $\{\delta_j : j \notin \mathcal{H}\}$, not only on $\Delta_{\mathcal{H}}$.
For hotspot-localized attacks, where perturbations are restricted to $\mathcal{H}$, we use a local-effect approximation: the dominant logit changes occur inside $\mathcal{H}$, while non-hotspot logits remain approximately unchanged, i.e., $\delta_j\approx 0$ for $j\notin\mathcal{H}$.
Under this approximation, softmax normalization implies that any non-hotspot region $R \subset \mathcal{V} \setminus \mathcal{H}$ is expected to gain attention in proportion to its pre-attack attention:
\begin{equation}
\bar a'_R - \bar a_R
\;\approx\;
\bar a_R
\frac{\Delta_{\mathcal{H}}}{1-m_{\mathcal{H}}}
\;\propto\;
\bar a_R .
\label{eq:main_redistribution}
\end{equation}
A full derivation is provided in Appendix~\ref{app:softmax_derivation}.

\rev{We next examine whether regions with higher pre-attack mean attention exhibit larger attention gains, as suggested by Equation~\ref{eq:main_redistribution}. We observe a positive correlation under hotspot-localized attacks (Pearson $r=0.409$, $p=2\times10^{-70}$), whereas the random-region control shows no significant correlation ($r\approx 0$, $p>0.5$) (Appendix~\ref{app:total_pred}). These observations motivate using pre-attack attention to order subsequent update regions.}

\begin{figure*}[!t]
    \centering
    \includegraphics[width=0.95\textwidth]{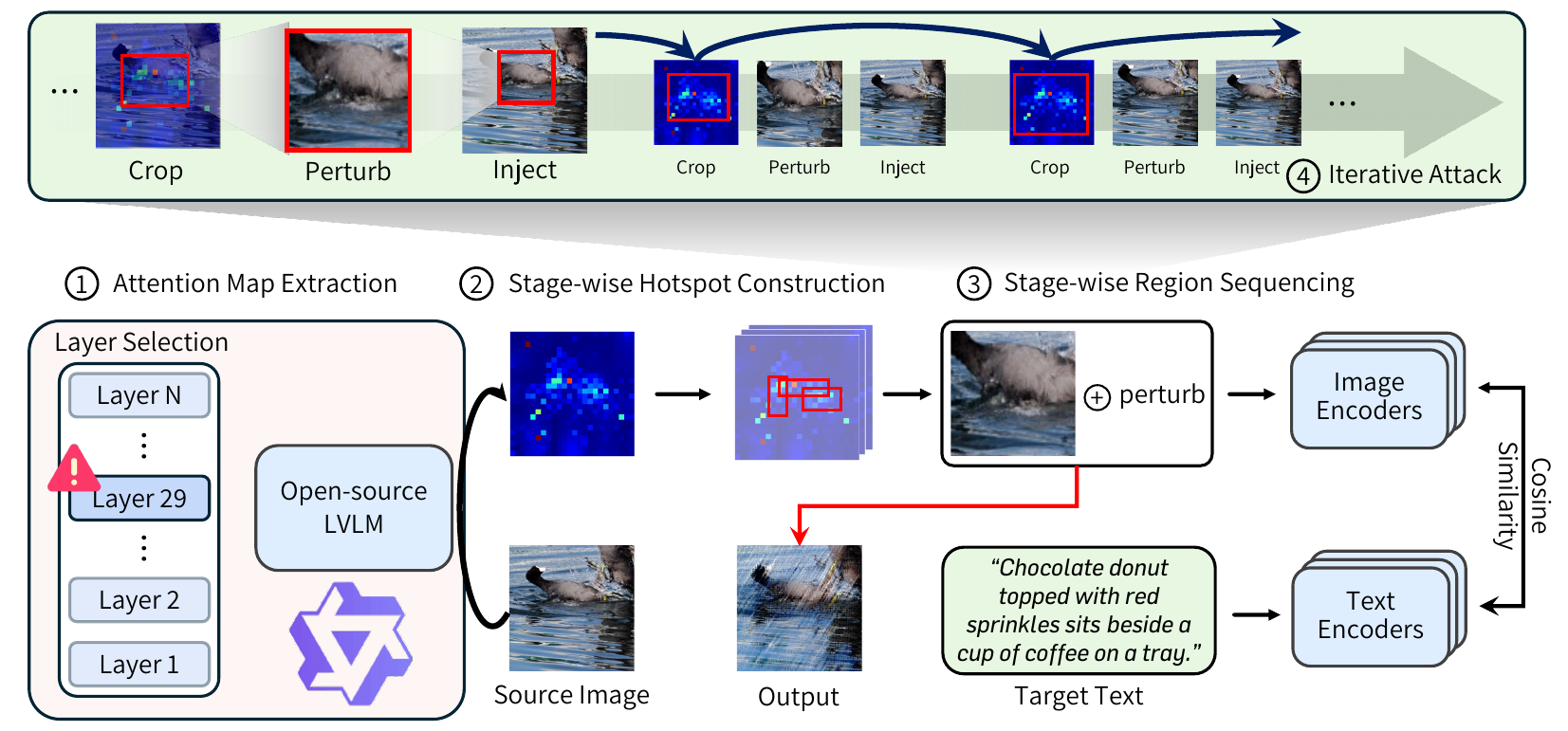}
    \caption{Overview of \system{}. Given a source image, \system{} extracts a cross-modal attention map from an open-source LVLM and constructs a stage-wise sequence of high-attention hotspots. During optimization, perturbation updates are restricted to the scheduled hotspot while the adversarial image remains within the global $L_\infty$ constraint. The sequence progressively expands from the initial hotspot to next-salient regions precomputed from the fixed pre-attack attention map, allocating the perturbation budget according to the attention-based principles in Section~\ref{sec:motivation}.} 
    \label{fig:main_figure}
    \vspace{-10pt}
\end{figure*}

\section{Stage-wise Attention-Guided Region Sequencing}
\label{sec:method}

Building on the two principles in Section~\ref{sec:motivation}, \system{} implements attention-guided region sequencing as a lightweight component on top of standard transfer-based attacks. 
It leaves the surrogate objective unchanged and requires no access to the target model's parameters, gradients, or attention maps. 
Instead, \system{} determines which attention-guided regions receive localized updates and in what order. 
Figure~\ref{fig:main_figure} illustrates the overall procedure, with the full algorithm provided in Appendix~\ref{appendix:algorithm}.

\noindent\textbf{Attention Map Extraction.}
Before optimization, \system{} extracts a single cross-modal attention map from an open-source LVLM by prompting it to caption the source image $x_{\mathrm{orig}}$. 
The map is kept fixed throughout the attack and used only to construct the region sequence. 
Following the sensitivity analysis in Section~\ref{subsec:motiv_where}, we use Qwen3-VL-8B~\citep{bai2025qwen3vltechnicalreport} as the default extractor because its regional attention scores show the strongest correlation with adversarial loss sensitivity.
\rev{Extraction details are provided in Appendix~\ref{appendix:attention_extraction}, and sensitivity to the extractor model and layer is evaluated in Appendix~\ref{appendix:attention_extractor_mdoel_ablation}.}

\noindent\textbf{Stage-wise Hotspot Construction.}
Given the attention map, \system{} constructs an ordered hotspot sequence before optimization. 
The attack is divided into $N$ stages, where stage $n$ is associated with an area ratio $a_n = n/N$. 
For each stage, we identify the top-$a_n$ fraction of visual patches ranked by attention score and select the top-$k$ hotspot regions with the highest regional attention scores. 
To avoid redundant updates, we enforce an intersection-over-union (IoU) constraint so that the IoU between any pair of selected hotspots is below $\tau$. 
This produces $k \times N$ hotspots that expand from the most concentrated attention region to broader next-salient regions.
Unless otherwise specified, we use $N=10$, $k=3$, and $\tau=0.3$; Appendix~\ref{app:nk_ablation} ablates these choices.

\noindent\textbf{Stage-wise Region Sequencing.}
Let $E$ denote the total number of attack iterations. 
\system{} assigns $E/(kN)$ iterations to each hotspot in the ordered sequence.
During each iteration block, perturbation updates are restricted to the corresponding hotspot, while the adversarial image remains constrained within the $L_\infty$ ball around $x_{\mathrm{orig}}$. 
Within each hotspot, we sample local subregions for update, following the localized attack primitive used in prior transfer attacks~\citep{li2025a}.
Unlike stochastic localization over the full image, this sequence begins with highly sensitive hotspots and moves the update support toward next-salient regions precomputed from the fixed pre-attack attention map, as motivated by the redistribution analysis.

\noindent\textbf{Optimization.}
Given the scheduled hotspot, \system{} samples a local subregion and maximizes the average cosine similarity between the adversarial image embedding and the target-text embedding over an ensemble of surrogate CLIP models:
\begin{equation}
\mathcal{L}(x_{\mathrm{adv}}^{r}, y_{\mathrm{targ}})
=
\frac{1}{S}
\sum_{i=1}^{S}
\cos\!\left(
f_{\mathrm{img}}^{i}(x_{\mathrm{adv}}^{r}),
f_{\mathrm{txt}}^{i}(y_{\mathrm{targ}})
\right).
\end{equation}
At each iteration, the gradient update is applied only within the scheduled hotspot, and the adversarial image is projected back to the global $L_\infty$ ball around $x_{\mathrm{orig}}$. 
Thus, \system{} modifies only the region sequence for localized perturbation updates while remaining compatible with existing transfer-based objectives.

\section{Experiments}
\label{sec:experiment}

\subsection{Experimental Setups}
\label{subsec:experiment_setup}

\noindent\textbf{Datasets.}
We use 1,000 source images sampled from the NIPS 2017 Adversarial Attacks and Defenses Competition dataset~\citep{alexey2017nips}, resized to $224 \times 224 \times 3$, following previous works~\citep{jia2025adversarial}.
To construct target texts, we randomly sample 1,000 images from the MSCOCO validation set and generate captions using Qwen3-VL-8B-Instruct.

\begin{table*}[t]
\caption{Comparison with state-of-the-art adversarial attack methods in terms of attack effectiveness. We report attack success rate (ASR) and average similarity (AvgSim) across \rev{fourteen} LVLMs. \rev{Each cell reports ASR / AvgSim.} Bold indicates the best performance for each metric.}
\centering
\fontsize{8.8pt}{10.0pt}\selectfont
\renewcommand{\arraystretch}{1.22}
\setlength{\tabcolsep}{2pt}
\begin{tabular}{@{}p{0.12\textwidth}*{7}{>{\centering\arraybackslash\leavevmode}p{\dimexpr(0.88\textwidth-28pt)/7\relax}}@{}}
\toprule
\multicolumn{8}{l}{\textbf{Closed-source}} \\
\midrule
\raisebox{0.35\baselineskip}{\textbf{Method}} & \raisebox{-1pt}[\height][\depth]{\textbf{\shortstack{Gemini\\2.5 Flash}}} & \textbf{\shortstack{Gemini\\3 Pro}} & \rev{\raisebox{-1pt}[\height][\depth]{\textbf{\shortstack{Gemini\\3.8 Flash}}}} & \raisebox{-1pt}[\height][\depth]{\textbf{\shortstack{Grok\\4 Fast}}} & \raisebox{0.35\baselineskip}{\textbf{GPT-4.1}} & \raisebox{0.35\baselineskip}{\textbf{GPT-5 mini}} & \rev{\raisebox{-1pt}[\height][\depth]{\textbf{\shortstack{GPT-5.6\\Luna}}}} \\
\midrule
X-Transfer & 0.01 / 0.06 & 0.00 / 0.06 & \rev{0.00} / \rev{0.06} & 0.01 / 0.06 & 0.01 / 0.07 & 0.01 / 0.06 & \rev{0.00} / \rev{0.07} \\
AnyAttack & 0.04 / 0.10 & 0.02 / 0.07 & \rev{0.04} / \rev{0.10} & 0.07 / 0.13 & 0.07 / 0.13 & 0.06 / 0.12 & \rev{0.04} / \rev{0.12} \\
M-Attack & 0.36 / 0.37 & 0.22 / 0.26 & \rev{0.31} / \rev{0.35} & 0.44 / 0.43 & 0.57 / 0.54 & 0.35 / 0.36 & \rev{0.09} / \rev{0.17} \\
FOA-Attack & 0.35 / 0.37 & 0.25 / 0.28 & \rev{0.35} / \rev{0.37} & 0.45 / 0.44 & 0.62 / 0.56 & 0.36 / 0.37 & \rev{0.08} / \rev{0.17} \\
\rowcolor{gray!15}
\textbf{\system{}} & \rev{\textbf{0.49}} / \textbf{0.49} & \textbf{0.35} / \textbf{0.37} & \rev{\textbf{0.43}} / \rev{\textbf{0.43}} & \rev{\textbf{0.59} / \textbf{0.53}} & \rev{\textbf{0.67}} / \textbf{0.60} & \rev{\textbf{0.41}} / \rev{\textbf{0.41}} & \rev{\textbf{0.12}} / \rev{\textbf{0.19}} \\
\bottomrule
\end{tabular}
\par\vspace{0.5em}
\begin{tabular}{@{}p{0.12\textwidth}*{7}{>{\centering\arraybackslash\leavevmode}p{\dimexpr(0.88\textwidth-28pt)/7\relax}}@{}}
\toprule
\multicolumn{8}{l}{\textbf{Open-source}} \\
\midrule
\raisebox{0.35\baselineskip}{\textbf{Method}} & \textbf{\shortstack{LLaVA-1.5\\7B}} & \textbf{\shortstack{Gemma\\3-4B}} & \textbf{\shortstack{Llama 4\\Maverick}} & \rev{\raisebox{-1pt}[\height][\depth]{\textbf{\shortstack{DeepSeek\\V4.1 Flash}}}} & \raisebox{-1pt}[\height][\depth]{\textbf{\shortstack{Qwen3-VL\\30B}}} & \raisebox{-1pt}[\height][\depth]{\textbf{\shortstack{Qwen3-VL\\235B}}} & \rev{\raisebox{-1pt}[\height][\depth]{\textbf{\shortstack{Qwen3.8\\27B}}}} \\
\midrule
X-Transfer & 0.01 / 0.05 & 0.00 / 0.04 & 0.01 / 0.06 & \rev{0.01} / \rev{0.06} & 0.01 / 0.06 & 0.01 / 0.07 & \rev{0.00} / \rev{0.06} \\
AnyAttack & 0.07 / 0.12 & 0.03 / 0.08 & 0.07 / 0.12 & \rev{0.06} / \rev{0.11} & 0.07 / 0.13 & 0.07 / 0.13 & \rev{0.07} / \rev{0.12} \\
M-Attack & 0.62 / 0.53 & 0.17 / 0.21 & 0.41 / 0.40 & \rev{0.40} / \rev{0.40} & 0.52 / 0.48 & 0.57 / 0.53 & \rev{0.42} / \rev{0.42} \\
FOA-Attack & 0.65 / 0.56 & 0.18 / 0.22 & 0.42 / 0.41 & \rev{0.43} / \rev{0.42} & 0.53 / 0.50 & 0.62 / 0.55 & \rev{0.42} / \rev{0.42} \\
\rowcolor{gray!15}
\textbf{\system{}} & \textbf{0.67} / \textbf{0.58} & \textbf{0.25} / \textbf{0.28} & \rev{\textbf{0.44}} / \textbf{0.43} & \rev{\textbf{0.56}} / \rev{\textbf{0.51}} & \textbf{0.57} / \textbf{0.52} & \rev{\textbf{0.63}} / \textbf{0.58} & \rev{\textbf{0.47}} / \rev{\textbf{0.44}} \\
\bottomrule
\end{tabular}
\label{tab:main_asr}
\vspace{-2.5em}
\end{table*}
\normalsize

\noindent\textbf{Implementation Settings.}
We use three CLIP surrogate models: ViT-B/16, ViT-B/32, and ViT-g-14-laion2B-s12B-b42K. 
The same surrogates are used for M-Attack~\citep{li2025a}, FOA-Attack~\citep{jia2025adversarial}, and \system{}, isolating the effect of attention-guided region sequencing rather than surrogate capacity.
We set the perturbation budget to $\epsilon=16/255$, the step size to $1/255$, and run all attacks for \rev{300 iterations}.
For \system{}, we use $N=10$, $k=3$, and $\tau=0.3$ unless otherwise specified.
\rev{We extract attention from layer 29 of Qwen3-VL-8B. Appendix~\ref{appendix:attention_extractor_mdoel_ablation} examines the choice of extractor, model-family alignment, and extraction layer.}

\noindent\textbf{Target Models.} 
\rev{The main comparison evaluates fourteen target LVLMs}, comprising \rev{seven} closed-source models (Gemini 2.5 Flash, \rev{Gemini 3 Pro}, \rev{Gemini 3.8 Flash}, \rev{Grok 4 Fast}, GPT-4.1, \rev{GPT-5 mini}, and \rev{GPT-5.6 Luna}) and \rev{seven} open-source models (\rev{LLaVA-1.5-7B}, Gemma 3-4B, Llama 4 Maverick, \rev{DeepSeek V4.1 Flash}, \rev{Qwen3-VL-30B}, Qwen3-VL-235B-A22B-Instruct, and \rev{Qwen3.8 27B}).

\noindent\textbf{Evaluation Metrics.}
Following prior work~\citep{li2025a, jia2025adversarial}, we employ an LLM-as-a-judge framework~\citep{zheng2023judging}.
Specifically, gpt-oss-20b~\citep{agarwal2025gpt} computes similarity scores between captions generated from attacked images and their target texts.
We report Attack Success Rate (ASR), where an attack is successful if the similarity score exceeds 0.5, and Average Similarity Score (AvgSim)~\citep{jia2025adversarial}.
We additionally report ASR under alternative thresholds (0.3, 0.7, 0.8) and KMR~\citep{li2025a} in Appendix~\ref{appendix:asr_thresholds} and Appendix~\ref{app:kmr}.
\rev{We also evaluate captions using Claude Sonnet 4.6 and Gemini 3.5 Flash as additional LLM judges (Appendix~\ref{app:independent_judges}).}

To evaluate imperceptibility, we report normalized $\ell_1$ and $\ell_2$ distances, PSNR~\citep{huynh2008scope}, SSIM~\citep{wang2004image}, LPIPS~\citep{zhang2018unreasonable}, and BRISQUE~\citep{mittal2012no}.
The $\ell_1$ and $\ell_2$ distances measure pixel-level distortion, PSNR and SSIM quantify reconstruction fidelity, LPIPS captures perceptual feature distance, and BRISQUE evaluates the naturalness of the attacked image itself.

\subsection{Comparison Results}

Table~\ref{tab:main_asr} reports ASR and AvgSim across the \rev{fourteen} target LVLMs, while Table~\ref{tab:imperceptibility} summarizes imperceptibility metrics.
Localized attacks such as M-Attack~\citep{li2025a} and FOA-Attack~\citep{jia2025adversarial} substantially outperform earlier baselines, confirming the effectiveness of localized perturbation updates.
However, these methods still rely on stochastic localization over the entire image. 
In contrast, \system{} uses attention-guided region sequencing to determine which regions receive updates and how the perturbation budget is allocated across them.

\rev{\system{} achieves the highest ASR and AvgSim among the compared methods on all fourteen targets in Table~\ref{tab:main_asr}.}
\rev{Compared to the strongest baseline for each target, \system{} yields relative ASR improvements of 40.0\% on Gemini 3 Pro and 30.2\% on DeepSeek V4.1 Flash, with consistent gains on the remaining targets.}
Notably, \system{} extracts attention once from a single open-source LVLM before optimization, yet transfers effectively across all targets, suggesting that the attention signal captures transferable vulnerability patterns rather than model-specific artifacts.

\rev{\system{} also achieves the best imperceptibility across all six metrics in Table~\ref{tab:imperceptibility}. Its higher ASR is accompanied by lower pixel distortion and better perceptual quality. Section~\ref{subsec:analysis_ablations} examines the perturbation budget used to obtain these gains and the contributions of region selection and ordering.}

\Needspace{13\baselineskip}
\begin{wraptable}[11]{r}{0.53\linewidth}
\vspace{-12pt}
\caption{
Comparison of imperceptibility metrics. Lower is better for $\ell_1$, $\ell_2$, LPIPS, and BRISQUE (BRQ); higher is better for SSIM and PSNR.
}
\label{tab:imperceptibility}
\vspace{3pt}
\centering
\fontsize{7.5pt}{8.6pt}\selectfont
\renewcommand{\arraystretch}{1.10}
\setlength{\tabcolsep}{2.7pt}
\resizebox{\linewidth}{!}{%
\begin{tabular}{lcccccc}
\toprule
\textbf{Method} 
& $\ell_1\downarrow$ 
& $\ell_2\downarrow$ 
& SSIM$\uparrow$ 
& PSNR$\uparrow$ 
& LPIPS$\downarrow$ 
& BRQ$\downarrow$ \\
\midrule
X-Transfer  & 0.2026 & 0.2351 & 0.3456 & 12.73 & 0.3548 & 52.29 \\
AnyAttack   & 0.0530 & 0.0561 & 0.6667 & 25.03 & 0.3354 & 27.50 \\
M-Attack    & 0.0317 & 0.0372 & 0.7417 & 28.59 & 0.2279 & 26.84 \\
FOA-Attack  & 0.0321 & 0.0376 & 0.7387 & 28.51 & 0.2288 & 26.71 \\
\rowcolor{gray!15}
\textbf{\system{}} 
& \textbf{0.0269} 
& \textbf{0.0330} 
& \textbf{0.7823} 
& \textbf{29.65} 
& \textbf{0.1965} 
& \textbf{22.51} \\
\bottomrule
\end{tabular}
}
\vspace{-12pt}
\end{wraptable}

\leavevmode\rev{To examine whether the observed gains depend on the choice of evaluator, we reassess captions from five selected target models using Claude Sonnet 4.6 and Gemini 3.5 Flash. \system{} achieves the highest ASR among the compared methods under both judges. Although absolute ASRs vary across evaluators, this consistent advantage indicates that \system{}'s effectiveness is not specific to the original gpt-oss-20b judge. Detailed results are provided in Appendix~\ref{app:independent_judges}.}

Although \system{} is not specifically designed as a defense-robust attack, Appendix~\ref{app:defense_eot} evaluates common input transformations~\citep{jpeg, gaussian_blur} and DPS~\citep{dps}, and shows that a simple EOT~\citep{eot} variant can be applied orthogonally to improve robustness.

\subsection{\rev{Analysis and Ablations}}
\label{subsec:analysis_ablations}

\begin{figure}[!t]
    \centering
    \begin{minipage}[t]{0.47\linewidth}
        \centering
        \includegraphics[width=\linewidth]{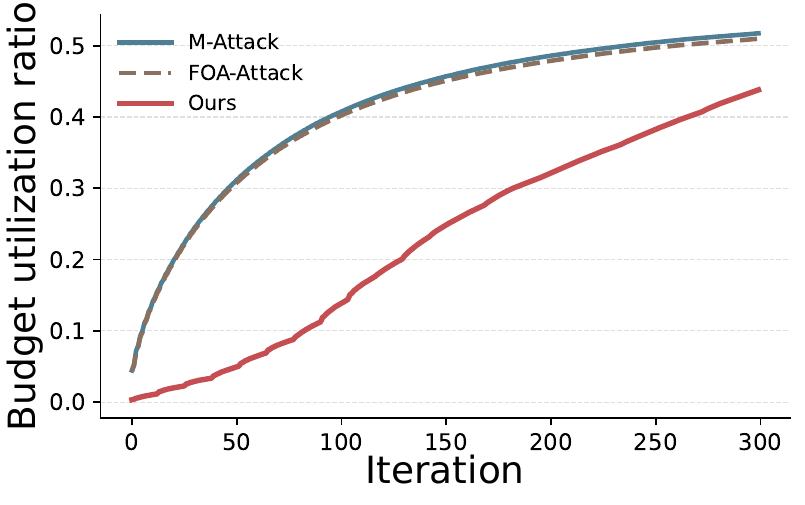}
        \caption{\rev{Budget utilization over attack iterations. \system{} reaches a lower final ratio while achieving higher ASR.}}
        \label{fig:exp_budget_saturation}
    \end{minipage}\hfill
    \begin{minipage}[t]{0.49\linewidth}
        \centering
        \includegraphics[width=\linewidth]{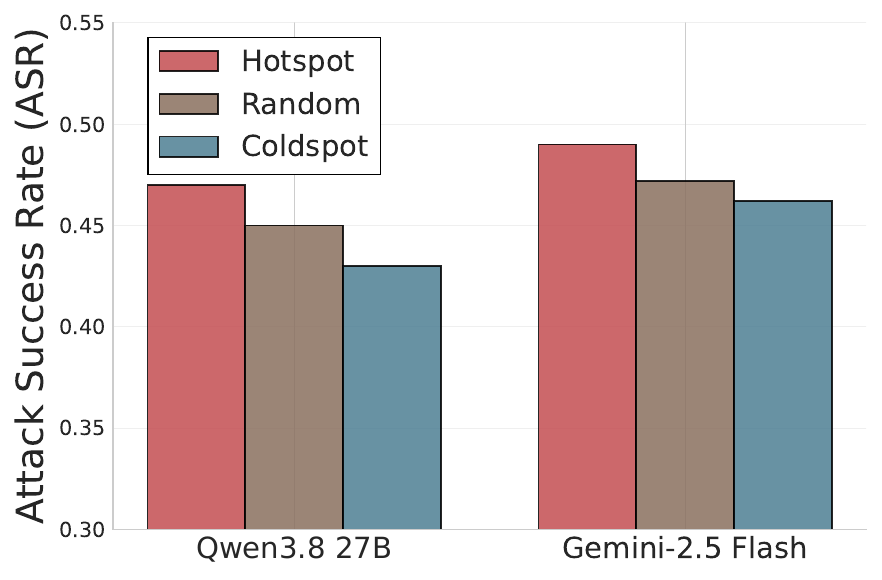}
        \caption{\rev{Region selection under the same stage-wise schedule. High-attention hotspots yield higher ASR than random regions and coldspots.}}
        \label{fig:exp_hotspot_vs_coldspot}
    \end{minipage}
    \vspace{-10pt}
\end{figure}

\begin{revision}
\noindent\textbf{Perturbation Efficiency under Budget Constraints.}
To relate attack effectiveness to perturbation magnitude, we define the \emph{budget utilization ratio} as follows.
\begin{equation}
\label{eq:budget_saturation}
\text{budget utilization ratio}
=\frac{1}{|\mathcal{P}|}\sum_{p\in\mathcal{P}}
\min\!\left(\frac{|\delta_p|}{\epsilon},1\right),
\end{equation}
where $\delta_p=x_p^{\mathrm{adv}}-x_p^{\mathrm{orig}}$, $\mathcal{P}$ is the set of all pixel-channel entries in the image, and $\epsilon$ is the perturbation budget.

Under the $L_\infty$ constraint, this ratio equals the mean absolute perturbation divided by $\epsilon$.
Figure~\ref{fig:exp_budget_saturation} shows that M-Attack and FOA-Attack rapidly increase budget utilization, whereas \system{} increases it more gradually and reaches a lower final value.
Together with its higher ASR, this result indicates more effective use of the perturbation budget.
Lower utilization directly corresponds to smaller average pixel distortion, consistent with the improved imperceptibility metrics in Table~\ref{tab:imperceptibility}.
Additionally, \system{} achieves higher ASR despite producing the smallest mean clean-to-adversarial cosine distance in CLIP feature space among the compared methods (Appendix~\ref{app:feature_distance}).

\noindent\textbf{Region Selection: Hotspot vs. Random vs. Coldspot.}
To examine the contribution of region selection, we compare high-attention hotspots, randomly sampled regions, and low-attention coldspots under the same stage-wise schedule.
We evaluate Gemini 2.5 Flash and Qwen3.8 27B as closed-source and open-source targets, respectively.
Figure~\ref{fig:exp_hotspot_vs_coldspot} shows that hotspot selection achieves higher ASR than random regions and coldspots on both targets.
On Gemini 2.5 Flash, hotspot selection achieves 49\% ASR, compared with 47.2\% for random regions and 46.2\% for coldspots.
On Qwen3.8 27B, hotspot selection achieves 47\% ASR, compared with 45\% for random regions and 43\% for coldspots, improving ASR by 2 and 4 percentage points, respectively.

These results show that \system{} benefits from prioritizing high-attention regions rather than merely imposing a stage-wise update schedule.
Consistent with the sensitivity analysis in Section~\ref{subsec:motiv_where}, allocating updates to regions identified as more sensitive by pre-attack attention improves attack effectiveness.

\noindent\textbf{Controlled Ablation on Region Ordering.}
To isolate the effect of ordering, we fix the pre-attack hotspot set, region sizes, perturbation support, and total number of updates, and change only the update schedule.
We compare the proposed sequence with its reverse, a random permutation, and a cyclic schedule that visits the next hotspot at every iteration.
As shown in Table~\ref{tab:region_ordering}, \system{} achieves the highest ASR on three targets and ties with reverse and cyclic ordering on Gemini 2.5 Flash.
For example, on Qwen3-VL-235B, ASR is 0.63 for \system{} and 0.46--0.51 for the alternative schedules.
Appendix~\ref{app:ordering_details} describes the schedules.

Averaged across the four targets, ASR increases from reverse to random, cyclic, and the proposed schedule. Reverse follows the opposite order, random permutes it, and cyclic follows the proposed order but repeatedly returns to smaller regions.
\system{} achieves the highest average ASR by following this order in contiguous stages, progressing toward larger regions without restarting from smaller ones.

Together, the region-selection and ordering experiments distinguish two design choices.
The former shows the benefit of allocating updates to high-attention regions.
The latter supports the proposed update order on three of the four targets and on average, even when the hotspot set and total update budget are held fixed.
\system{} therefore benefits from both selecting sensitive regions and organizing their updates into the proposed stage-wise sequence.
\end{revision}

\begin{table}[t]
\centering\color{black}\small
\caption{\rev{Controlled region-ordering ablation. All schedules use the same hotspot set and total update budget. ASR is evaluated at threshold 0.5. Average denotes the mean across the four targets.}}
\label{tab:region_ordering}
\vspace{4pt}
\setlength{\tabcolsep}{4pt}
\begin{tabular}{lccccc}
\toprule
Schedule & Gemini 2.5 Flash & Llama 4 & Qwen3-VL-235B & GPT-5 mini & Average \\
\midrule
Reverse & \textbf{0.49} & 0.32 & 0.46 & 0.33 & 0.40 \\
Random & 0.47 & 0.35 & 0.48 & 0.32 & 0.41 \\
Cyclic & \textbf{0.49} & 0.37 & 0.51 & 0.36 & 0.43 \\
\rowcolor{gray!15}
\system{} & \textbf{0.49} & \textbf{0.44} & \textbf{0.63} & \textbf{0.41} & \textbf{0.49} \\
\bottomrule
\end{tabular}
\vspace{-11pt}
\end{table}

\begin{table}[t]
\centering\color{black}
\caption{\rev{Fixed vs. dynamic attention on nine targets. ASR is evaluated at threshold 0.5. Dynamic recomputes attention using the current adversarial image and a regenerated caption. Average ASR is calculated across all nine targets. Runtime is the time required to attack one image. Bold marks higher ASR or lower runtime.}}
\label{tab:fixed_dynamic}
\vspace{4pt}
\fontsize{8pt}{9.4pt}\selectfont
\renewcommand{\arraystretch}{1.18}
\setlength{\tabcolsep}{1.5pt}
\resizebox{\linewidth}{!}{%
\begin{tabular}{@{}l*{9}{c}>{\columncolor{gray!15}}c>{\columncolor{gray!15}}c@{}}
\toprule
Method & \shortstack{Gemini\\2.5 Flash} & \shortstack{Gemini\\3.8 Flash} & GPT-4.1 & \shortstack{GPT-5\\mini} & \shortstack{GPT-5.6\\Luna} & \shortstack{Llama 4\\Maverick} & \shortstack{DeepSeek\\V4.1 Flash} & \shortstack{Qwen3-VL\\235B} & \shortstack{Qwen3.8\\27B} & \textbf{\shortstack{Average\\ASR}} & \textbf{\shortstack{Runtime\\(s/image)}} \\
\midrule
\system{} (fixed) & 0.49 & 0.43 & \textbf{0.67} & \textbf{0.41} & \textbf{0.12} & 0.44 & \textbf{0.56} & 0.63 & \textbf{0.47} & 0.47 & \textbf{80.81} \\
Dynamic & \textbf{0.50} & \textbf{0.45} & 0.66 & 0.40 & \textbf{0.12} & \textbf{0.48} & 0.55 & \textbf{0.66} & \textbf{0.47} & \textbf{0.48} & 324.46 \\
\bottomrule
\end{tabular}%
}
\vspace{-10pt}
\end{table}

\begin{revision}
\noindent\textbf{Fixed vs. Dynamic Attention Maps.}
Section~\ref{subsec:motiv_where} shows that high-attention regions are more sensitive to localized attacks, while Section~\ref{subsec:motiv_when} characterizes attention redistribution under a local-effect approximation.
These findings motivate constructing the attack schedule from a single pre-attack attention map, avoiding repeated attention extraction during optimization.
To assess the benefit of updating this schedule, we compare \system{} with Dynamic, which redefines the hotspots at each stage using the current adversarial image.

Table~\ref{tab:fixed_dynamic} reports the comparison on nine targets.
Dynamic achieves higher ASR on four targets, ties with \system{} on two, and yields lower ASR on three.
Mean ASR is 0.47 for \system{} and 0.48 for Dynamic, indicating a small average advantage for dynamic recomputation.
However, repeated attention extraction increases the runtime per image from 80.81 to 324.46 seconds, approximately a fourfold increase.
These results support the efficiency of a precomputed schedule motivated by attention redistribution, retaining competitive attack success while avoiding the cost of repeatedly updating the attention map.

\end{revision}
\begin{figure}[!t]
    \centering
    \includegraphics[width=0.75\linewidth]{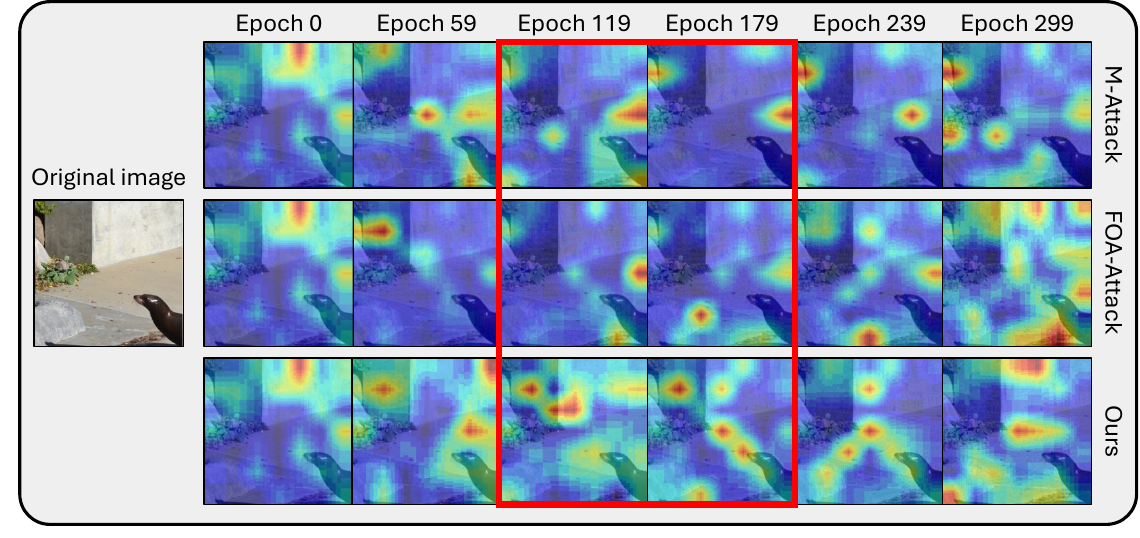}
    \caption{\rev{Attention maps extracted every 60 attack iterations for analysis only. The examples illustrate changes around initially dominant and subsequent salient regions.}}
    \label{fig:attention_changes_qual}
    \vspace{-10pt}
\end{figure}

\begin{wraptable}{r}{0.53\linewidth}
\vspace{-13.5pt}
\caption{\rev{VQAv2 soft accuracy on 100 image--question pairs. Lower accuracy after attack indicates a stronger attack.}}
\label{tab:vqa_transfer}
\vspace{-5pt}
\centering\begingroup
\fontsize{7.5pt}{8.6pt}\selectfont
\renewcommand{\arraystretch}{1.15}
\setlength{\tabcolsep}{2.7pt}
\resizebox{\linewidth}{!}{%
\begin{tabular}{lccccc}
\toprule
Method & Gemini & Llama 4 & Qwen & GPT-5 mini & Avg. \\
\midrule
Clean & 0.770 & 0.507 & 0.837 & 0.740 & 0.713 \\
M-Attack & 0.507 & 0.293 & 0.547 & 0.530 & 0.469 \\
FOA-Attack & 0.537 & \textbf{0.280} & 0.550 & 0.523 & 0.473 \\
\rowcolor{gray!15}
\system{} & \textbf{0.483} & 0.297 & \textbf{0.493} & \textbf{0.497} & \textbf{0.443} \\
\bottomrule
\end{tabular}}
\par\smallskip
\raggedright Gemini and Qwen denote Gemini 2.5 Flash and Qwen3-VL-235B, respectively.
\vspace{-12pt}
\par\endgroup
\end{wraptable}

\begin{revision}
\noindent\textbf{Evaluation Beyond Image Captioning.}
We further evaluate whether \system{} can disrupt visual question answering using 100 image--question pairs from VQAv2.
Each target model answers the same questions using clean and attacked images, allowing us to measure how image perturbations degrade answer accuracy.
Table~\ref{tab:vqa_transfer} shows that \system{} reduces average accuracy across four targets from 71.34\% on clean images to 44.25\%, compared with 46.92\% for M-Attack and 47.25\% for FOA-Attack.
It yields the lowest accuracy on three of the four targets and the lowest average accuracy overall.
These results show that \system{} also effectively disrupts question answering beyond the captioning setting.
\end{revision}

\begin{revision}
\noindent\textbf{Visualization of Attention Map Changes.}
Finally, we qualitatively examine how attention evolves during optimization.
Figure~\ref{fig:attention_changes_qual} shows attention maps extracted every 60 iterations for M-Attack, FOA-Attack, and \system{}.
These maps are collected only for analysis and do not modify \system{}'s fixed attention map or update schedule.
In the displayed examples, \system{} shows a clearer progression from initially dominant regions toward subsequent salient regions, whereas the baselines exhibit more diffuse or fluctuating attention patterns.
Attention around the initial hotspots decreases as other salient regions become more prominent.
These observations complement the quantitative redistribution analysis in Section~\ref{subsec:motiv_when} and illustrate how attention evolves during the proposed stage-wise attack.
\end{revision}

\FloatBarrier
\section{Conclusion}

We formulated targeted adversarial attacks on LVLMs as a perturbation budget allocation problem under an $L_\infty$ constraint. 
\rev{Our attention-based analysis links regional attention to adversarial sensitivity and provides a conditional explanation of redistribution following hotspot perturbations.}
Based on these principles, we proposed \system{}, a stage-wise attention-guided region-sequencing framework that uses a fixed attention map from an open-source LVLM to allocate perturbation updates over precomputed high-attention hotspots. 
Across \rev{fourteen} closed-source and open-source LVLMs, \system{} achieves state-of-the-art attack success rates while also obtaining the best imperceptibility across all evaluated distortion and perceptual metrics. \rev{Controlled ablations demonstrate the benefit of the proposed region order, and fixed attention achieves similar ASR to dynamic recomputation at substantially lower computational cost. These results support region sequencing as an effective way to allocate localized updates under finite perturbation budgets.}

\clearpage
\renewcommand{\acksection}{\section*{Acknowledgments}}
\begin{ack}
This work was supported by the Institute of Information \& Communications Technology Planning \& Evaluation (IITP) grant funded by the Korean government (MSIT) (No. RS-2019-II190075 Artificial Intelligence Graduate School Support Program (KAIST), 5\%; No. RS-2024-00457882, National AI Research Lab Project, 5\%; No. RS-2026-25522672, Development of Unified Reasoning Technology Mimicking Human Cognition for Hierarchical Understanding and Unbounded Problem Solving, 40\%; No. (IITP-2026)-RS-2026-25614738, AI Star Fellowship Support Program, 25\%; No. IITP-2026-RS-2026-25548323, Artificial Intelligence Innovation Human Resources Development, 25\%).

The authors declare no competing interests.
\end{ack}

\clearpage
\appendix
\etocdepthtag.toc{appendix}
\begingroup
\hypersetup{hidelinks}
\color{black}
\renewcommand{\rev}[1]{#1}
\vspace*{0pt}
{\fontsize{16}{19}\selectfont\bfseries Appendix\par}
\vspace{12pt}
\hrule height 0.6pt
\vspace{8pt}
\etocsettagdepth{main}{none}
\etocsettagdepth{appendix}{subsection}
\etocsettocstyle{}{}
\etocstandardlines
\makeatletter
\renewcommand*{\l@section}[2]{%
  \vspace{5pt}%
  {\bfseries\@dottedtocline{1}{0em}{2em}{#1}{#2}}}
\renewcommand*{\l@subsection}[2]{%
  \@dottedtocline{2}{2em}{2.7em}{#1}{#2}}
\makeatother
\fontsize{9}{10.5}\selectfont
\begingroup
\renewcommand{\color}[2][]{}
\tableofcontents
\endgroup
\clearpage
\endgroup

\section{Related Work}
\label{appendix:related_work}

\subsection{Large Vision--Language Models}
Large Vision--Language Models (LVLMs) extend large language models by incorporating visual modalities, enabling joint reasoning over images and text. Trained on large-scale image--text data, LVLMs have demonstrated strong performance on multimodal tasks such as image captioning~\citep{li2022blip}, visual question answering~\citep{kuang2025natural}, and cross-modal reasoning~\citep{chen2024spatialvlm}. Recent advances in LVLMs have been driven by both open-source and commercial models. Representative open-source systems include BLIP-2~\citep{li2023blip}, LLaVA~\citep{NEURIPS2023_6dcf277e}, Qwen~\citep{bai2023qwen, team2024qwen2, yang2025qwen3}, and Gemma~\citep{team2024gemma, team2025gemma}, while commercial models such as GPT~\citep{radford2018improving, achiam2023gpt} and Gemini~\citep{team2023gemini, comanici2025gemini} further extend multimodal capabilities.

\subsection{Adversarial Attacks on Vision--Language Models}
Despite their strong performance, LVLMs remain vulnerable due to structural gaps introduced by the integration of visual encoders with pre-trained language models~\citep{zhao2025jailbreaking}. Motivated by this vulnerability, recent work has explored adversarial attacks on LVLMs from various perspectives. X-Transfer~\citep{huang2025xtransfer} demonstrates the existence of highly transferable adversarial perturbations, while AnyAttack~\citep{zhang2025anyattack} enables large-scale targeted attacks without requiring explicit target labels. More recent studies highlight the effectiveness of spatially localized perturbations. M-Attack~\citep{li2025a} and FOA-Attack~\citep{jia2025adversarial} show that random cropping and feature alignment can significantly improve attack success, but rely on stochastic cropping strategies and do not explicitly prioritize vulnerable regions, often leading to inefficient use of the perturbation budget.

\subsection{Cross-modal Attention in Vision--Language Models}
Recent studies have explored cross-modal attention in vision--language models as a tool for interpreting model behavior and diagnosing their limitations~\citep{kaduri2025s, neo2025towards}. Such analyses examine how visual information is attended to during multimodal reasoning and provide insights into model decision processes. Prior work has shown that attention patterns can help explain failures in spatial reasoning~\citep{chen2025why} or be leveraged to mitigate hallucinations~\citep{jiang2024interpreting}. Overall, existing approaches primarily treat attention as an interpretability or correction mechanism, aiming to analyze model behavior. In contrast, our work repurposes cross-modal attention as a proactive attack signal: we extract attention maps from open-source models to identify vulnerable regions within an image and leverage this signal to guide adversarial perturbations, enabling more effective and budget-efficient attacks against LVLMs.

\section{\rev{Limitations and Future Work}}
\label{appendix:limitations}
\begin{revision}
\noindent\textbf{Dependence on the attention extractor.}
\system{} uses cross-modal attention from an open-source LVLM as a proxy for regional adversarial sensitivity.
Performance varies with the extractor and layer, and attention may miss regions relevant to the target model.
Future work could combine layers or extractors and select informative maps without target-model access.

\noindent\textbf{Scope of the redistribution analysis.}
Our softmax analysis explains attention reallocation under a local-effect approximation, whereas the measurements show nonzero logit changes outside the attacked hotspot.
It therefore motivates the region order rather than predicting the full attention trajectory during optimization.
Future analysis could incorporate changes in text queries and visual keys to characterize when the initial ranking remains informative.

\end{revision}

\newpage
\section{Algorithm}
\label{appendix:algorithm}

\begin{algorithm}[h]
   \caption{Stage-wise Attention-Guided Attack (\system{})}
   \label{algorithm}
   \begin{algorithmic}[1]
      \Require Source image $x_{\mathrm{orig}}\in[0,1]$, target text $y_{\mathrm{targ}}$,
      attention extractor $\mathcal{F}$, surrogate CLIP models
      $\{(f^{i}_{\mathrm{img}}, f^{i}_{\mathrm{txt}})\}_{i=1}^{S}$,
      perturbation budget $\epsilon$, step size $\eta$, total iterations $E$, number of stages $N$,
      maximum hotspots per stage $k$, IoU threshold $\tau$

      \State $M \gets \mathcal{F}(x_{\mathrm{orig}}, \text{``Describe this image.''})$
      \Comment{Extract attention once}
      \State $(\mathcal{R}_1,\ldots,\mathcal{R}_J) \gets \textsc{BuildStageWiseHotspots}(M,N,k,\tau)$
      \Comment{Flatten in stage order}
      \State $b \gets \lfloor E/J\rfloor$, $\delta \gets 0$
      \Comment{Require $E\geq J$}
      \For{$j=1$ to $J$}
         \State $E_j \gets b + \mathbf{1}[j=J]\,(E-bJ)$
         \Comment{Assign remainder to the last hotspot}
         \For{$e=1$ to $E_j$}
            \State $r \gets \textsc{RandomResizedCrop}(\mathcal{R}_j,\ \mathrm{scale}=[0.5,0.9])$
            \State $\mathcal{L} \gets \frac{1}{S}\sum_{i=1}^{S}
            \cos\!\left(f^{i}_{\mathrm{img}}((x_{\mathrm{orig}}+\delta)^r),
            f^{i}_{\mathrm{txt}}(y_{\mathrm{targ}})\right)$
            \State $\delta \gets \operatorname{clip}_{[-\epsilon,\epsilon]}
            \left(\delta+\eta\,\operatorname{sign}(\nabla_{\delta}\mathcal{L})\right)$
         \EndFor
      \EndFor
      \State \Return $\operatorname{clip}_{[0,1]}(x_{\mathrm{orig}}+\delta)$
   \end{algorithmic}
\end{algorithm}

\noindent\textsc{BuildStageWiseHotspots} selects up to $k$ distinct regions per stage, first applying the IoU cap and then filling any remaining slots with the highest-scoring unused candidates even if they overlap. If fewer candidates exist, fewer regions are returned. Thus, $J$ is the realized schedule length, and $\sum_{j=1}^{J}E_j=E$. Pixel values, $\epsilon$, and $\eta$ are expressed on the $[0,1]$ scale above; the implementation uses the equivalent $[0,255]$ scale.

\section{Softmax Redistribution: Derivation and Empirical Verification}
\label{app:softmax_verification}

In Section~\ref{subsec:motiv_when}, we introduced a prediction under a local-effect approximation relating attention changes in non-hotspot regions to the attention mass vacated from the attacked hotspot.
This appendix provides the full derivation of the softmax redistribution model and the experimental setup supporting the empirical claims summarized in the main text.

\subsection{Derivation of the Softmax Redistribution Model}
\label{app:softmax_derivation}

At a selected transformer layer and generated text token $t$, let $z_i$ denote the attention logit assigned to visual token $i \in \mathcal{V}$. 
We write the normalized token-level attention weight as
\begin{equation}
a_i
=
\mathrm{softmax}(z)_i
=
\frac{\exp(z_i)}
{\sum_{k \in \mathcal{V}} \exp(z_k)}
=
\frac{\exp(z_i)}{Z},
\label{eq:appendix_softmax_attention}
\end{equation}
where $z_i = \mathbf{q}_t^\top \mathbf{k}_i / \sqrt{d}$ is the attention logit, $\mathcal{V}$ is the set of visual tokens, and
\begin{equation}
Z = \sum_{k \in \mathcal{V}} \exp(z_k)
\end{equation}
is the softmax normalization constant. 
We normalize attention over the visual-token set, so $\sum_{i \in \mathcal{V}} a_i = 1$.
For a region $S \subset \mathcal{V}$, we define its total attention mass and mean attention score as
\begin{equation}
m_S = \sum_{i\in S} a_i,
\qquad
\bar a_S = \frac{1}{|S|}m_S.
\label{eq:appendix_region_notation}
\end{equation}

Let $\mathcal{H} \subset \mathcal{V}$ denote the attacked hotspot.
After the adversarial update, suppose the logits change from $z_i$ to $z_i + \delta_i$, yielding post-attack token-level attention
\begin{equation}
a'_i
=
\frac{\exp(z_i + \delta_i)}
{Z'},
\qquad
Z'
=
\sum_{k \in \mathcal{V}} \exp(z_k + \delta_k).
\label{eq:appendix_post_attention}
\end{equation}
We use $m'_S$ and $\bar a'_S$ to denote the corresponding post-attack total mass and mean score for a region $S$.
Let the total hotspot attention mass decrease by
\begin{equation}
\Delta_{\mathcal{H}}
=
m_{\mathcal{H}} - m'_{\mathcal{H}} > 0.
\label{eq:appendix_delta_hotspot}
\end{equation}

We use a local-effect approximation: a perturbation restricted to $\mathcal{H}$ primarily changes the logits of visual tokens inside $\mathcal{H}$, while logits outside the hotspot remain approximately unchanged. 
That is, $\delta_j \approx 0$ for $j \notin \mathcal{H}$. 
For the derivation below, take $\delta_j = 0$ outside the hotspot; the resulting relation is approximate when these logits change slightly. Then, for any non-hotspot token $j \in \mathcal{V} \setminus \mathcal{H}$,
\begin{equation}
a'_j
=
\frac{\exp(z_j)}{Z'}
=
a_j \cdot \frac{Z}{Z'}.
\label{eq:appendix_nonhotspot_ratio}
\end{equation}

Summing Equation~\ref{eq:appendix_nonhotspot_ratio} over any non-hotspot region $R \subset \mathcal{V} \setminus \mathcal{H}$ gives
\begin{equation}
m'_R
=
m_R \cdot \frac{Z}{Z'}.
\label{eq:appendix_region_mass_ratio}
\end{equation}

To express the normalization ratio in terms of the hotspot attention change, we sum Equation~\ref{eq:appendix_nonhotspot_ratio} over the full complement $\mathcal{V} \setminus \mathcal{H}$:
\begin{equation}
1 - m'_{\mathcal{H}}
=
(1 - m_{\mathcal{H}})
\cdot
\frac{Z}{Z'}.
\label{eq:appendix_complement_ratio}
\end{equation}
Since $m'_{\mathcal{H}} = m_{\mathcal{H}} - \Delta_{\mathcal{H}}$, Equation~\ref{eq:appendix_complement_ratio} gives
\begin{equation}
\frac{Z}{Z'}
=
1
+
\frac{\Delta_{\mathcal{H}}}
{1 - m_{\mathcal{H}}}.
\label{eq:appendix_z_ratio}
\end{equation}
Substituting Equation~\ref{eq:appendix_z_ratio} into Equation~\ref{eq:appendix_region_mass_ratio}, we obtain
\begin{equation}
m'_R - m_R
=
m_R
\cdot
\frac{\Delta_{\mathcal{H}}}
{1 - m_{\mathcal{H}}},
\qquad
R \subset \mathcal{V} \setminus \mathcal{H}.
\label{eq:appendix_redistribution_mass}
\end{equation}

Equation~\ref{eq:appendix_redistribution_mass} is exact when non-hotspot logits are unchanged and approximate when they are only weakly affected. 
Dividing both sides by $|R|$ yields the corresponding rule for the mean regional attention score:
\begin{equation}
\bar a'_R - \bar a_R 
\;\approx\;
\bar a_R \cdot \frac{\Delta_{\mathcal{H}}}{1 - m_{\mathcal{H}}}
\;\propto\;
\bar a_R,
\qquad
R \subset \mathcal{V} \setminus \mathcal{H}.
\label{eq:appendix_redistribution_mean}
\end{equation}
Thus, among non-hotspot regions of comparable size, the expected attention gain is proportional to the pre-attack regional mean attention. 
\rev{This motivates prioritizing next-salient regions when the local-effect approximation holds.}

\subsection{Empirical Verification Setup}
\label{app:softmax_empirical_setup}

\textbf{Data.} 
We use 1000 image--target text pairs drawn from the same evaluation set as the main experiments: source images from the NIPS 2017 Adversarial Attacks and Defenses Competition dataset, resized to $224 \times 224 \times 3$, and target captions generated by Qwen3-VL-8B-Instruct on MSCOCO validation images.

\textbf{Attention extraction.} 
Pre- and post-attack attention maps are extracted from Qwen3-VL-8B at layer 29, which shows the highest attention--loss correlation in Figure~\ref{fig:appendix_layer_wise_correlation}. 
The attention maps are head-averaged and normalized to sum to 1 over the $24 \times 24$ patch grid.

\textbf{Attack procedure.} 
We apply a localized attack restricted to the hotspot $\mathcal{H}$, using the same $L_\infty$ budget and step size as in the main experiments. 
The attack is run to convergence on the hotspot region.

\textbf{Random-region attacks.} 
We replace $\mathcal{H}$ with a random region $\mathcal{R}_{\mathrm{rand}}$ sampled to be disjoint from $\mathcal{H}$. 
The attack procedure is otherwise identical.

\textbf{Regions of interest.} 
We partition each image into three regions:
\begin{itemize}[leftmargin=1.5em, itemsep=0pt, topsep=2pt]
    \item $\mathcal{R}_1 = \mathcal{H}$: the attacked hotspot, defined as the top-10\% highest-attention region.
    \item $\mathcal{R}_2 = \mathrm{Top20} \setminus \mathcal{H}$: the next-salient region, defined as the top-20\% highest-attention patches with the hotspot $\mathcal{R}_1$ removed.
    \item $\mathcal{R}_3 = \mathcal{V} \setminus \mathrm{Top20}$: the coldspot region.
\end{itemize}
For each evaluated region $R$, we compute its pre-attack mean attention $\bar a_R$ and observed attention change $\Delta\bar a_R=\bar a'_R-\bar a_R$.
Motivated by the proportional relationship in Equation~\ref{eq:appendix_redistribution_mean}, we measure the Pearson correlation between these quantities across evaluated region instances.

\subsection{Direction of Mass Flow}
\label{app:direction}

Table~\ref{tab:direction} reports the pre- and post-attack region-averaged attention under the hotspot attack. 
The attacked hotspot loses attention ($\Delta\bar a_{\mathcal{R}_1} = -3.40 \times 10^{-3}$), while both $\mathcal{R}_2$ and $\mathcal{R}_3$ gain attention. 
\rev{The increase in mean attention in the next-salient region $\mathcal{R}_2$ is $2.47\times$ that in the coldspot $\mathcal{R}_3$, consistent with the trend motivated by the redistribution analysis.}

\begin{table}[h]
\centering
\caption{Pre- and post-attack region means under the hotspot attack.}
\label{tab:direction}
\begin{tabular}{lccc}
\toprule
Region & $\bar a$ (pre) & $\bar a'$ (post) & $\Delta \bar a$ \\
\midrule
$\mathcal{R}_1$ (hotspot)       & $6.84 \times 10^{-3}$ & $3.45 \times 10^{-3}$ & $-3.40 \times 10^{-3}$ \\
$\mathcal{R}_2$ (next-salient)  & $1.74 \times 10^{-3}$ & $2.49 \times 10^{-3}$ & $+7.52 \times 10^{-4}$ \\
$\mathcal{R}_3$ (coldspot)      & $1.11 \times 10^{-3}$ & $1.42 \times 10^{-3}$ & $+3.05 \times 10^{-4}$ \\
\bottomrule
\end{tabular}
\end{table}

\subsection{Correlation between Pre-Attack Attention and Attention Change}
\label{app:total_pred}

We examine whether regions with higher pre-attack attention tend to exhibit larger attention gains under hotspot-localized attacks.
Table~\ref{tab:total_corr} reports the Pearson correlation between pre-attack regional mean attention $\bar a_R$ and observed attention change $\Delta\bar a_R$ across evaluated region instances, under hotspot attacks and the random-region control.

\begin{table}[h]
\centering
\caption{Pearson correlation between pre-attack regional mean attention and observed attention change across evaluated region instances.}
\label{tab:total_corr}
\begin{tabular}{lcc}
\toprule
Condition & Pearson $r$ & $p$-value \\
\midrule
Hotspot attack & $\mathbf{+0.409}$ & $2 \times 10^{-70}$  \\
Random control & $\approx 0.00$ & $> 0.5$ \\
\bottomrule
\end{tabular}
\end{table}

Pre-attack regional mean attention is positively correlated with observed attention change for hotspot-localized attacks (Pearson $r=0.409$, $p=2\times10^{-70}$), but shows no significant correlation for the random-region control.

Together with the larger increase in mean attention in the next-salient region than in the coldspot region (Table~\ref{tab:direction}), this result supports using pre-attack attention to guide region sequencing.

\FloatBarrier
\section{Experimental Details}

\subsection{Cross-Modal Attention Map Extraction}
\label{appendix:attention_extraction}

To analyze how generated text grounds in the visual input, we extract cross-modal attention weights from intermediate transformer layers of a vision--language model (VLM). Let the model consist of $L$ transformer layers. During autoregressive generation, each layer $\ell \in \{1,\ldots,L\}$ produces attention matrices at every decoding step.

Formally, let the VLM generate a token sequence $(x_1, \ldots, x_{T_0}, y_1, \ldots, y_T)$, where $\{x_i\}_{i=1}^{T_0}$ are prompt tokens, including vision tokens, and $\{y_t\}_{t=1}^{T}$ are generated text tokens. Let $\mathcal{V} \subset \{1,\ldots,T_0\}$ denote the contiguous index set of vision tokens corresponding to image patches, with $|\mathcal{V}| = H \times W$ for a spatial grid of resolution $(H,W)$.

At decoding step $t$, let $A^{\ell}_{t\rightarrow i}\in[0,1]$ denote the attention weight assigned to token $i$ at layer $\ell$ when generating $y_t$, after averaging over attention heads.

\paragraph{Token-to-vision attention projection.}
For each selected generated token $y_t$ with $t \in \mathcal{T} \subseteq \{1,\ldots,T\}$, \rev{where $\mathcal{T}$ denotes the generated content-word tokens selected for aggregation, with all generated tokens used if the selection is empty,} we extract its attention mass over vision tokens $\mathcal{V}$:
\[
\mathbf{a}_t
=
\left( A^{\ell}_{t\rightarrow i} \right)_{i \in \mathcal{V}}
\in \mathbb{R}^{HW}.
\]
We reshape $\mathbf{a}_t$ into a spatial attention map $\mathbf{A}_t \in \mathbb{R}^{H \times W}$. Then, we normalize the spatial attention map $\mathbf{A}_t$ so that the total attention mass over all spatial locations sums to one:
\[
\tilde{\mathbf{A}}_t^{h,w}
=
\frac{\mathbf{A}_t^{h,w}}
{\sum_{h'=1}^{H}\sum_{w'=1}^{W} \mathbf{A}_t^{h',w'}},
\qquad
\forall\, h \in \{1,\dots,H\},\; w \in \{1,\dots,W\}.
\]

\paragraph{Aggregation across tokens.}
The final attention map is obtained by averaging over all valid generated tokens:
\[
\mathbf{A}
=
\frac{1}{|\mathcal{T}|}
\sum_{t \in \mathcal{T}} \tilde{\mathbf{A}}_t.
\]

\subsection{Correlation Analysis: Cross-Modal Attention and Adversarial Loss Sensitivity}
\label{appendix:correlation_exp}

\begin{figure*}[!t]
    \centering
    \includegraphics[width=0.98\textwidth]{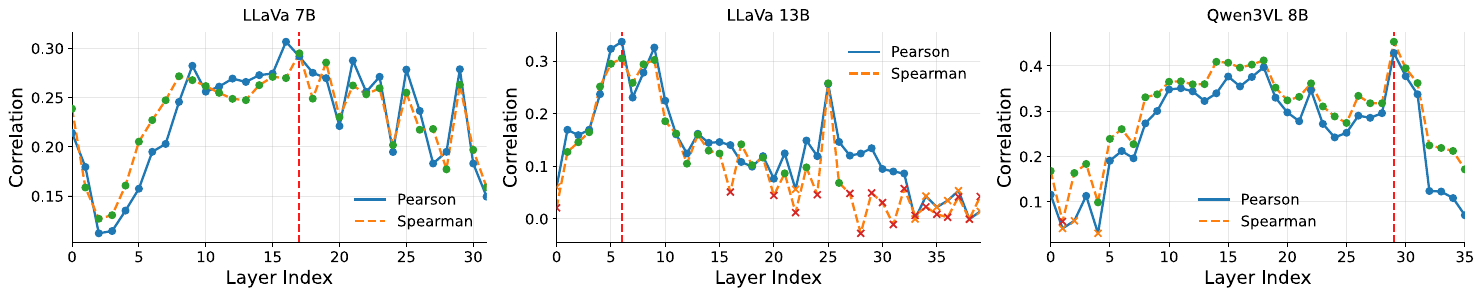}
    \caption{Layer-wise correlation between regional attention scores and adversarial loss sensitivity. Markers with crosses indicate layers where the correlation is not statistically significant ($p > 0.05$).}
    \label{fig:appendix_layer_wise_correlation}
\end{figure*}

We conduct a correlation analysis on three open-source LVLMs: LLaVA-7B, LLaVA-13B, and Qwen3-VL-8B. The goal of this experiment is to examine the relationship between regional cross-modal attention scores and adversarial loss sensitivity across different transformer layers.

Prior studies on attention maps in vision--language models~\citep{chen2025why, kaduri2025s} suggest that different layers exhibit distinct behaviors, with certain layers more effectively attending to semantically meaningful image regions. Motivated by this observation, we perform a layer-wise correlation analysis rather than aggregating attention signals across layers.

For the correlation analysis, we construct a controlled evaluation set by selecting 10 source images and 5 target texts. We form all Cartesian combinations between images and texts, resulting in 50 image--text pairs. For each pair, we randomly sample 20 spatial crop regions and apply a single localized adversarial update to each region, yielding a total of 1,000 samples. For every sample, we record the average cross-modal attention score within the cropped region and the corresponding change in adversarial loss, which are then used to compute layer-wise Pearson and Spearman correlations.

Figure~\ref{fig:appendix_layer_wise_correlation} reports the Pearson and Spearman correlation coefficients computed at each layer for the three models. Layers marked with crosses indicate cases where the correlation is not statistically significant, i.e., the corresponding $p$-value exceeds 0.05. Across all models, we observe that most layers exhibit a positive correlation, indicating that regions receiving higher attention tend to be more sensitive to adversarial perturbations.

Notably, the highest correlation is achieved at the 17th layer for LLaVA-7B, the 6th layer for LLaVA-13B, and the 29th layer for Qwen3-VL-8B. Based on these results, we extract attention maps from all attention heads at the layer with the highest correlation for each model and average them to obtain a single attention map, which is used throughout all experiments.

\subsection{Negative Control: Hotspot Suppression Is Not Sufficient}
\label{app:negative_control}

We compare hotspot-localized attacks with random-region attacks to test whether reducing the initial hotspot attention is sufficient for predictable redistribution. 
We partition visual tokens into three disjoint regions based on the initial attention map:
\[
\mathcal{R}_{\mathrm{hot}} = \mathcal{H}_{0.1}, 
\qquad
\mathcal{R}_{\mathrm{next}} = \mathcal{H}_{0.2}\setminus\mathcal{H}_{0.1},
\qquad
\mathcal{R}_{\mathrm{cold}} = \mathcal{V}\setminus\mathcal{H}_{0.2}.
\]
Here, $\mathcal{H}_{0.1}$ denotes the top-10\% highest-attention region, $\mathcal{H}_{0.2}$ denotes the top-20\% highest-attention region, $\mathcal{R}_{\mathrm{next}}$ denotes the next-salient region after excluding the initial hotspot, and $\mathcal{R}_{\mathrm{cold}}$ denotes the remaining coldspot region.

\begin{figure*}[!t]
    \centering
    \begin{minipage}[t]{0.41\textwidth}
        \centering
        \includegraphics[width=\linewidth]{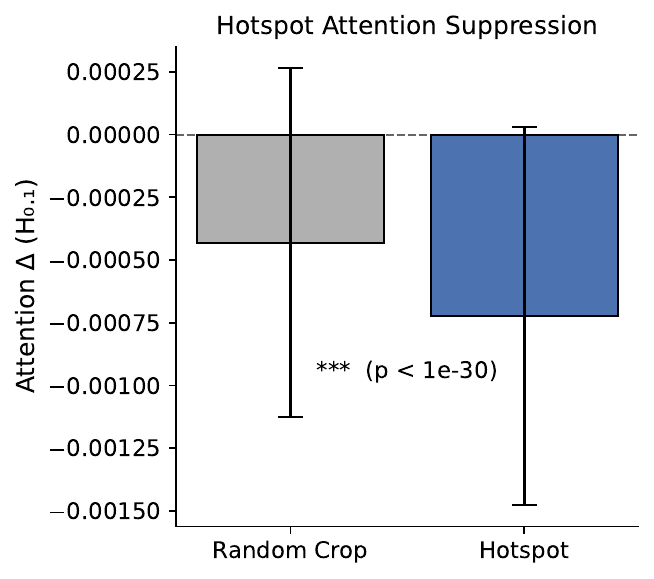}
        \caption{Both random-region and hotspot-localized attacks reduce the attention mass of the initial top-10\% hotspot $\mathcal{R}_{\mathrm{hot}}$.}
        \label{fig:app_hotspot_suppression}
    \end{minipage}
    \hspace{0.02\textwidth}
    \begin{minipage}[t]{0.54\textwidth}
        \centering
        \includegraphics[width=\linewidth]{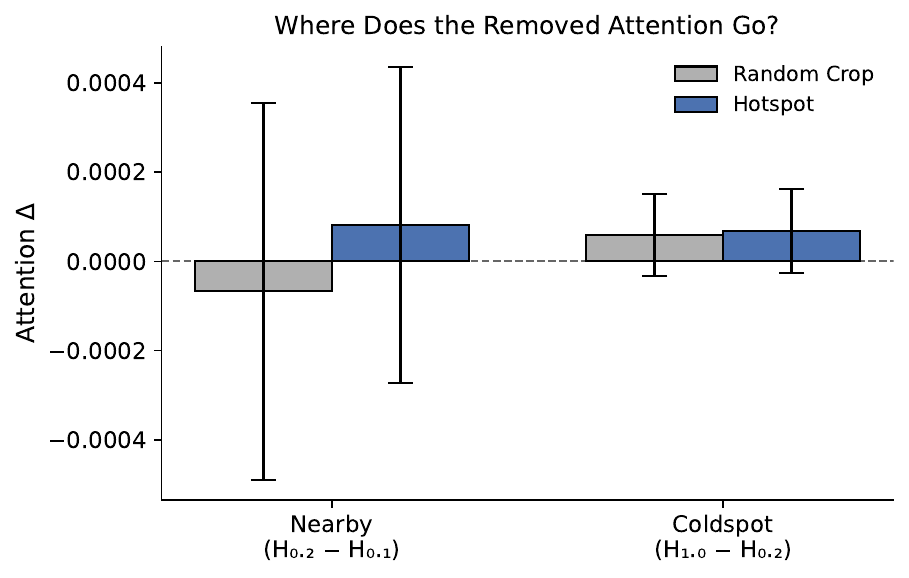}
        \caption{Random-region attacks redistribute attention mainly toward coldspot regions $\mathcal{R}_{\mathrm{cold}}$, producing diffuse and unstructured shifts. In contrast, hotspot-localized attacks preferentially increase attention in the next-salient region $\mathcal{R}_{\mathrm{next}}$, providing a structured signal for stage-wise region sequencing.}
        \label{fig:app_redistribution_direction}
    \end{minipage}
\end{figure*}

Figure~\ref{fig:app_hotspot_suppression} shows that both random-region and hotspot-localized attacks decrease the attention of $\mathcal{R}_{\mathrm{hot}}$. 
Figure~\ref{fig:app_redistribution_direction} shows the difference in redistribution direction: random-region attacks primarily shift attention toward coldspot regions, whereas hotspot-localized attacks preferentially shift attention toward the next-salient region $\mathcal{R}_{\mathrm{next}}$. 
This confirms that predictable sequencing requires hotspot-localized suppression, not merely hotspot attention reduction.

\section{Additional Results}
\label{app:additional_result}

\subsection{\rev{Implementation Details and Compute Resources}}
\label{app:compute_resources}

\rev{\textbf{Baseline implementations.} We reimplemented AnyAttack~\citep{zhang2025anyattack}, X-Transfer~\citep{huang2025xtransfer}, M-Attack~\citep{li2025a}, and FOA-Attack~\citep{jia2025adversarial} using their official codebases as references. We evaluated these implementations under our experimental protocol.}

All adversarial images were generated using NVIDIA RTX A5000 and NVIDIA RTX A6000 GPUs.
\rev{The main captioning comparison covers fourteen target LVLMs, comprising seven closed-source and seven open-source models (Table~\ref{tab:main_asr}). We construct 1,000 image--target pairs and run each attack for 300 iterations. Sample sizes for supplementary analyses are specified in their respective protocols.}
For LVLM evaluation, we queried target models through OpenRouter API calls to generate captions for attacked images. 
The LLM-as-a-judge evaluation was also conducted through OpenRouter API calls using the same evaluation pipeline.

\begin{revision}
\subsection{Controlled Region-Ordering Protocol}
\label{app:ordering_details}
The ordering experiment uses the same pre-attack attention map and hotspot list for all schedules, with $N=10$, $k=3$, and $\tau=0.3$.
We keep the surrogate ensemble, $L_\infty$ budget, and 300 attack iterations fixed.
The reverse schedule reverses the list, and the random schedule applies a permutation with a different seed.
These variants allocate a contiguous block of updates to each scheduled hotspot, as in \system{}.
The cyclic variant visits hotspots in the original list order but changes the active hotspot after every iteration, repeating the list until the iteration budget is reached.
All variants use the same local crop scale range of 0.5--0.9 within the scheduled hotspot.
Table~\ref{tab:region_ordering} reports the resulting ASRs.

\end{revision}

\begin{revision}
\subsection{CLIP Feature Distance}
\label{app:feature_distance}
To complement the pixel-level budget analysis, we measure cosine distance between clean and adversarial images in CLIP ViT-L/14 feature space.
We also compute its Pearson correlation with the attack similarity score on Gemini 2.5 Flash.
Table~\ref{tab:exp_feature_distance} shows a slightly smaller mean distance for \system{} than for M-Attack and FOA-Attack.
The correlation for \system{} is positive but weak ($r=0.118$, $p<0.001$).
These results show that higher attack success can accompany smaller average feature changes.
\end{revision}
\begin{table}[t!]
\centering
\small
\caption{\rev{Clean-to-adversarial CLIP feature distance and its Pearson correlation with attack similarity score on Gemini 2.5 Flash.}}
\label{tab:exp_feature_distance}
\begin{tabular}{lccc}
\toprule
Method & Feature Distance & Pearson $r$ & $p$-value \\
\midrule
M-Attack & 0.4922 & 0.041\textsuperscript{n.s.} & 0.191 \\
FOA-Attack & 0.4957 & 0.055\textsuperscript{n.s.} & 0.080 \\
\system{} & \textbf{0.4848} & \textbf{0.118}\textsuperscript{***} & $<$\textbf{0.001} \\
\bottomrule
\end{tabular}
\end{table}

\begin{revision}
\subsection{Evaluation with Independent LLM Judges}
\label{app:independent_judges}
We evaluate the generated captions of M-Attack, FOA-Attack, and \system{} using Claude Sonnet 4.6 and Gemini 3.5 Flash as additional judges.
We retain the caption--target comparison protocol and the ASR threshold of 0.5.
Table~\ref{tab:independent_judges} reports ASR as percentages.
\system{} ranks first on all five targets under both judges.
Absolute ASRs vary across judges, but the relative advantage of \system{} is consistent.
\end{revision}

\begin{table}[htbp]
\centering\small
\caption{\rev{ASR (\%) at threshold 0.5 under two additional LLM judges. Each judge evaluates generated captions against their target texts.}}
\label{tab:independent_judges}
\setlength{\tabcolsep}{5pt}
\begin{tabular}{lccccc}
\toprule
Method & Gemini 2.5 Flash & Llama 4 & Qwen3-VL-235B & GPT-5 mini & Grok 4 Fast \\
\midrule
\multicolumn{6}{l}{\textbf{Judge: Claude Sonnet 4.6}} \\
\midrule
M-Attack & 18.82 & 24.02 & 40.64 & 18.72 & 28.13 \\
FOA-Attack & 19.72 & 25.63 & 40.84 & 19.82 & 27.23 \\
\rowcolor{gray!15}
\system{} & \textbf{26.33} & \textbf{26.33} & \textbf{45.75} & \textbf{23.42} & \textbf{38.54} \\
\midrule
\multicolumn{6}{l}{\textbf{Judge: Gemini 3.5 Flash}} \\
\midrule
M-Attack & 25.23 & 30.83 & 46.15 & 23.72 & 33.73 \\
FOA-Attack & 26.73 & 31.43 & 47.95 & 25.63 & 34.63 \\
\rowcolor{gray!15}
\system{} & \textbf{35.84} & \textbf{32.43} & \textbf{51.55} & \textbf{28.73} & \textbf{47.15} \\
\bottomrule
\end{tabular}
\end{table}

\subsection{ASR under Different Success Thresholds}
\label{appendix:asr_thresholds}

In the main experiments (Table~\ref{tab:main_asr}), we follow FOA-Attack~\citep{jia2025adversarial} and define an attack as successful if the LLM-judge similarity score exceeds 0.5. To verify that our conclusions are not specific to this choice of threshold, we additionally evaluate ASR under various thresholds of 0.3, 0.7, and 0.8 on four target models (Llama 4 Maverick, Grok 4 Fast, \rev{Gemini 2.5 Flash}, and GPT-5 mini). Tables~\ref{tab:appendix_asr_thr03}, \ref{tab:appendix_asr_thr07}, and~\ref{tab:appendix_asr_thr08} report the results.

Across all three thresholds, \system{} consistently achieves the highest ASR on every target model and the highest average ASR. Note that AvgSim is independent of the threshold by definition, so its value is identical across the three tables. \rev{Relative to the strongest baseline, the relative gain in average ASR is larger at thresholds 0.7 and 0.8 than at the standard threshold of 0.5. At threshold 0.8, \system{} achieves an average ASR of 0.20, compared with 0.14 for both M-Attack and FOA-Attack. These results show that the ASR advantage persists under stricter success criteria.}

\begin{table}[t!]
\centering
\small
\caption{ASR and AvgSim under success threshold $> 0.3$. \system{} achieves the highest ASR across all four target models and on average.}
\label{tab:appendix_asr_thr03}
\setlength{\tabcolsep}{4pt}
\resizebox{\textwidth}{!}{%
\begin{tabular}{lccccccccccc}
\toprule
& \multicolumn{2}{c}{Llama 4 Maverick} & \multicolumn{2}{c}{Grok 4 Fast} & \multicolumn{2}{c}{\rev{Gemini 2.5 Flash}} & \multicolumn{2}{c}{GPT-5 mini} & \multicolumn{2}{c}{Average} \\
\cmidrule(lr){2-3} \cmidrule(lr){4-5} \cmidrule(lr){6-7} \cmidrule(lr){8-9} \cmidrule(lr){10-11}
Method & ASR & AvgSim & ASR & AvgSim & ASR & AvgSim & ASR & AvgSim & ASR & AvgSim \\
\midrule
AnyAttack & 0.14 & 0.12 & 0.14 & 0.13 & 0.10 & 0.10 & \rev{0.15} & 0.12 & \rev{0.13} & 0.12 \\
X-Transfer & 0.03 & 0.06 & 0.03 & 0.06 & 0.04 & 0.06 & \rev{0.04} & 0.06 & \rev{0.04} & 0.06 \\
M-Attack & 0.58 & 0.40 & 0.64 & 0.43 & 0.54 & 0.37 & \rev{0.54} & 0.36 & \rev{0.58} & 0.39 \\
FOA-Attack & 0.61 & 0.41 & 0.65 & 0.44 & 0.55 & 0.37 & \rev{0.54} & 0.37 & \rev{0.59} & 0.40 \\
\rowcolor{gray!15} \system{} & \textbf{0.62} & \textbf{0.43} & \textbf{0.76} & \textbf{0.53} & \textbf{0.72} & \textbf{0.49} & \rev{\textbf{0.62}} & \textbf{0.41} & \rev{\textbf{0.68}} & \textbf{0.47} \\
\bottomrule
\end{tabular}%
}
\end{table}

\begin{table}[t!]
\centering
\small
\caption{ASR and AvgSim under success threshold $> 0.7$. \system{} maintains the highest ASR on every target and a clear margin on average.}
\label{tab:appendix_asr_thr07}
\setlength{\tabcolsep}{4pt}
\resizebox{\textwidth}{!}{%
\begin{tabular}{lccccccccccc}
\toprule
& \multicolumn{2}{c}{Llama 4 Maverick} & \multicolumn{2}{c}{Grok 4 Fast} & \multicolumn{2}{c}{\rev{Gemini 2.5 Flash}} & \multicolumn{2}{c}{GPT-5 mini} & \multicolumn{2}{c}{Average} \\
\cmidrule(lr){2-3} \cmidrule(lr){4-5} \cmidrule(lr){6-7} \cmidrule(lr){8-9} \cmidrule(lr){10-11}
Method & ASR & AvgSim & ASR & AvgSim & ASR & AvgSim & ASR & AvgSim & ASR & AvgSim \\
\midrule
AnyAttack & 0.04 & 0.12 & 0.03 & 0.13 & 0.02 & 0.10 & \rev{0.03} & 0.12 & \rev{0.03} & 0.12 \\
X-Transfer & 0.00 & 0.06 & 0.00 & 0.06 & 0.00 & 0.06 & \rev{0.00} & 0.06 & \rev{0.00} & 0.06 \\
M-Attack & 0.28 & 0.40 & 0.32 & 0.43 & 0.23 & 0.37 & \rev{0.25} & 0.36 & \rev{0.27} & 0.39 \\
FOA-Attack & 0.31 & 0.41 & 0.33 & 0.44 & 0.24 & 0.37 & \rev{0.25} & 0.37 & \rev{0.28} & 0.40 \\
\rowcolor{gray!15} \system{} & \textbf{0.33} & \textbf{0.43} & \textbf{0.45} & \textbf{0.53} & \textbf{0.37} & \textbf{0.49} & \rev{\textbf{0.30}} & \textbf{0.41} & \rev{\textbf{0.36}} & \textbf{0.47} \\
\bottomrule
\end{tabular}%
}
\end{table}

\begin{table}[t!]
\centering
\small
\caption{ASR and AvgSim under success threshold $> 0.8$. The relative gap between \system{} and the baselines becomes most pronounced under this strictest threshold.}
\label{tab:appendix_asr_thr08}
\setlength{\tabcolsep}{4pt}
\resizebox{\textwidth}{!}{%
\begin{tabular}{lccccccccccc}
\toprule
& \multicolumn{2}{c}{Llama 4 Maverick} & \multicolumn{2}{c}{Grok 4 Fast} & \multicolumn{2}{c}{\rev{Gemini 2.5 Flash}} & \multicolumn{2}{c}{GPT-5 mini} & \multicolumn{2}{c}{Average} \\
\cmidrule(lr){2-3} \cmidrule(lr){4-5} \cmidrule(lr){6-7} \cmidrule(lr){8-9} \cmidrule(lr){10-11}
Method & ASR & AvgSim & ASR & AvgSim & ASR & AvgSim & ASR & AvgSim & ASR & AvgSim \\
\midrule
AnyAttack & 0.01 & 0.12 & 0.01 & 0.13 & 0.00 & 0.10 & \rev{0.01} & 0.12 & \rev{0.01} & 0.12 \\
X-Transfer & 0.00 & 0.06 & 0.00 & 0.06 & 0.00 & 0.06 & \rev{0.00} & 0.06 & \rev{0.00} & 0.06 \\
M-Attack & 0.13 & 0.40 & 0.17 & 0.43 & 0.12 & 0.37 & \rev{0.13} & 0.36 & \rev{0.14} & 0.39 \\
FOA-Attack & 0.14 & 0.41 & 0.18 & 0.44 & 0.12 & 0.37 & \rev{0.13} & 0.37 & \rev{0.14} & 0.40 \\
\rowcolor{gray!15} \system{} & \textbf{0.16} & \textbf{0.43} & \textbf{0.27} & \textbf{0.53} & \textbf{0.19} & \textbf{0.49} & \rev{\textbf{0.16}} & \textbf{0.41} & \rev{\textbf{0.20}} & \textbf{0.47} \\
\bottomrule
\end{tabular}%
}
\end{table}

\subsection{KMRScore Evaluation}
\label{app:kmr}

In addition to the LLM-judge similarity score used in our main experiments, we evaluate \system{} under the KMRScore metric introduced by M-Attack~\citep{li2025a}. KMRScore pre-defines a set of keywords that describe the target text and measures the fraction of these keywords that appear in the caption generated from the attacked image, providing a more lexically grounded measure of attack success that complements the LLM-judge protocol. Following M-Attack~\citep{li2025a}, we report KMRScore$_a$, KMRScore$_b$, and KMRScore$_c$, which differ in the strictness of keyword matching, with KMRScore$_c$ being the most stringent. Tables~\ref{tab:appendix_kmr_part1} and~\ref{tab:appendix_kmr_part2} report results across four target models.

\system{} achieves the highest score on three out of four targets across all three KMRScore variants. The advantage is particularly pronounced on Grok 4 Fast and \rev{Gemini 2.5 Flash}, where \system{} surpasses the second-best baseline by sizable margins on KMRScore$_b$ (0.78 vs.\ 0.63 on Grok 4 Fast; 0.61 vs.\ 0.53 on \rev{Gemini 2.5 Flash}). On Llama 4 Maverick, \system{} performs comparably to FOA-Attack, suggesting that on this target the keyword-level signal saturates similarly for the two methods. Overall, the KMRScore results are consistent with the trends observed under the LLM-judge metric in our main experiments, indicating that the gains of \system{} are not specific to a single evaluation protocol.

\begin{table}[t!]
\centering
\small
\caption{Attack performance measured by KMRScore~\citep{li2025a} on Llama 4 Maverick and Grok 4 Fast.}
\label{tab:appendix_kmr_part1}
\setlength{\tabcolsep}{4pt}
\resizebox{\textwidth}{!}{%
\begin{tabular}{lcccccc}
\toprule
& \multicolumn{3}{c}{Llama 4 Maverick} & \multicolumn{3}{c}{Grok 4 Fast} \\
\cmidrule(lr){2-4} \cmidrule(lr){5-7}
Method & KMRScore$_a$ & KMRScore$_b$ & KMRScore$_c$ & KMRScore$_a$ & KMRScore$_b$ & KMRScore$_c$ \\
\midrule
AnyAttack & 0.36 & 0.14 & 0.02 & 0.40 & 0.23 & 0.03 \\
X-Transfer & 0.23 & 0.11 & 0.00 & 0.18 & 0.08 & 0.00 \\
M-Attack & 0.72 & 0.51 & 0.18 & 0.74 & 0.56 & 0.19 \\
FOA-Attack & \textbf{0.77} & \textbf{0.55} & \textbf{0.22} & 0.77 & 0.63 & 0.26 \\
\system{} & 0.76 & 0.54 & 0.19 & \textbf{0.90} & \textbf{0.78} & \textbf{0.36} \\
\bottomrule
\end{tabular}%
}
\end{table}

\begin{table}[t!]
\centering
\small
\caption{Attack performance measured by KMRScore~\citep{li2025a} on \rev{Gemini 2.5 Flash} and GPT-5 mini.}
\label{tab:appendix_kmr_part2}
\setlength{\tabcolsep}{4pt}
\resizebox{\textwidth}{!}{%
\begin{tabular}{lcccccc}
\toprule
& \multicolumn{3}{c}{\rev{Gemini 2.5 Flash}} & \multicolumn{3}{c}{GPT-5 mini} \\
\cmidrule(lr){2-4} \cmidrule(lr){5-7}
Method & KMRScore$_a$ & KMRScore$_b$ & KMRScore$_c$ & KMRScore$_a$ & KMRScore$_b$ & KMRScore$_c$ \\
\midrule
AnyAttack & 0.29 & 0.16 & 0.01 & 0.31 & 0.17 & 0.05 \\
X-Transfer & 0.21 & 0.10 & 0.00 & 0.24 & 0.10 & 0.01 \\
M-Attack & 0.68 & 0.45 & 0.12 & 0.67 & 0.51 & 0.16 \\
FOA-Attack & 0.70 & 0.53 & 0.11 & 0.67 & 0.48 & 0.14 \\
\system{} & \textbf{0.83} & \textbf{0.61} & \textbf{0.22} & \textbf{0.78} & \textbf{0.60} & \textbf{0.22} \\
\bottomrule
\end{tabular}%
}
\end{table}

\begin{figure}[!t]
\centering
\includegraphics[width=0.75\linewidth]{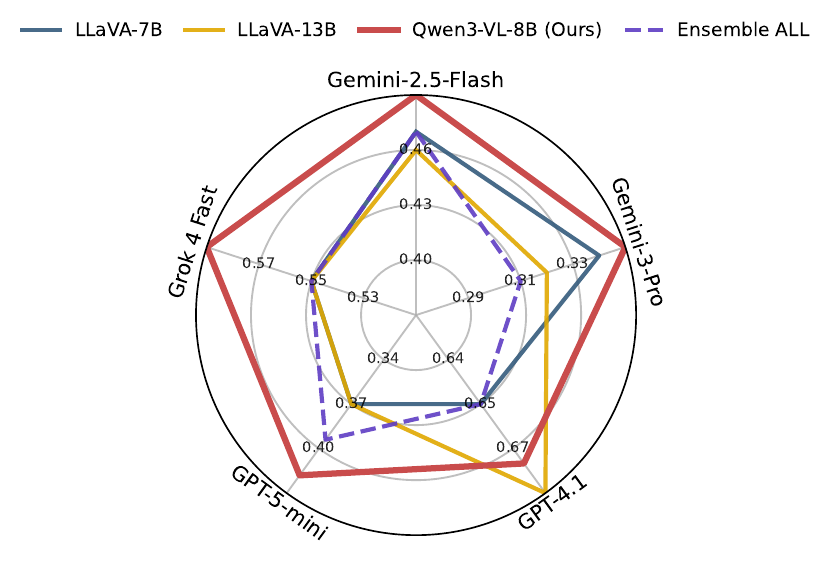}
\caption{\rev{Attention extractor comparison across closed-source targets. Qwen3-VL-8B achieves the highest reported ASR on four of the five targets.}}
\label{fig:ablation_attention_extractor}
\end{figure}

\FloatBarrier
\begin{revision}
\subsection{Attention Extractor, Model Family, and Layer Sensitivity}
\label{appendix:attention_extractor_mdoel_ablation}
\noindent\textbf{Extractor model.}
We compare attention maps from LLaVA-7B, LLaVA-13B, and Qwen3-VL-8B, along with an ensemble that averages the three maps.
Figure~\ref{fig:ablation_attention_extractor} reports results on Gemini 2.5 Flash, Gemini 3 Pro, GPT-4.1, GPT-5 mini, and Grok 4 Fast.
Qwen3-VL-8B yields the highest reported ASR on four of the five targets, while LLaVA-13B performs better on GPT-4.1. This supports Qwen3-VL-8B as a reliable default across the evaluated targets, although the best extractor depends on the target model.
Averaging the maps does not improve performance over Qwen3-VL-8B.

\noindent\textbf{Model-family alignment.}
For this model-family analysis, we use two targets from the main comparison, LLaVA-1.5-7B and Qwen3-VL-235B (Table~\ref{tab:appendix_attention_extractor_open}).
LLaVA extractors yield higher ASR on the LLaVA target, while Qwen3-VL-8B ties with LLaVA-13B on the Qwen target.
Cross-family extraction remains effective, and matching the target family does not consistently yield the highest ASR.

\noindent\textbf{Extraction layer.}
We fix the extractor to Qwen3-VL-8B and compare the default layer 29 with layer 17, which also has high attention--sensitivity correlation, and layer 4, which has lower correlation.
Table~\ref{tab:layer_sensitivity} holds the remaining attack settings fixed.
Layers 17 and 29 tie on Qwen3-VL-235B, while layer 29 is one percentage point higher on GPT-5 mini. Layer 4 yields lower ASR on both targets.
The similar performance of layers 17 and 29 supports using either high-correlation layer in this comparison. Layer 29 provides a reliable default, while the results do not identify a uniquely best layer across targets.
\end{revision}

\begin{table}[htbp]
\centering\small
\caption{\rev{Extractor-model ASR for the model-family analysis.}}
\label{tab:appendix_attention_extractor_open}
\begin{tabular}{lcc}
\toprule
Attention extractor & LLaVA-7B target & Qwen3-VL-235B target \\
\midrule
LLaVA-7B & 0.69 & 0.61 \\
LLaVA-13B & \textbf{0.71} & \textbf{0.63} \\
Qwen3-VL-8B & 0.67 & \textbf{0.63} \\
\bottomrule
\end{tabular}
\end{table}

\begin{table}[htbp]
\centering\small
\caption{\rev{Sensitivity to the Qwen3-VL-8B extraction layer. All other attack settings are fixed.}}
\label{tab:layer_sensitivity}
\begin{tabular}{lcc}
\toprule
Extraction layer & Qwen3-VL-235B & GPT-5 mini \\
\midrule
Layer 17 (high correlation) & \textbf{0.63} & 0.40 \\
Layer 4 (low correlation) & 0.58 & 0.36 \\
\rowcolor{gray!15}
Layer 29 (\system{}) & \textbf{0.63} & \textbf{0.41} \\
\bottomrule
\end{tabular}
\end{table}

\FloatBarrier
\subsection{Ablation on the Number of Stages and Hotspots}
\label{app:nk_ablation}

We ablate the number of stages $N$ and the number of hotspots per stage $k$ to evaluate the robustness of the stage-wise schedule.
When varying $N$, we fix $k=3$; when varying $k$, we fix $N=10$.
We evaluate on four representative target models: Llama 4 Maverick, Grok 4 Fast, \rev{Gemini 2.5 Flash}, and GPT-5 mini.
Table~\ref{tab:nk_ablation} reports ASR and AvgSim for each setting.

\rev{Among the tested configurations, the default setting $(N=10,k=3)$ achieves the highest average ASR and AvgSim across the four targets.
\rev{However, the best configuration varies by target. On Llama 4 Maverick, $N=30$ and $k=5$ yield ASRs of 0.45 and 0.46, respectively, compared with 0.44 for the default setting.}
\rev{Both alternatives have lower average ASR across the four targets, as do $N=5$ and $k=1$.}
These results support $(N=10,k=3)$ as a default that performs well across targets, while allowing that individual targets may benefit from different settings.}

\begin{table*}[t]
\centering
\caption{
Ablation on the number of stages $N$ and hotspots per stage $k$.
Each entry is reported as ASR / AvgSim.
The default setting $(N=10,k=3)$ achieves the best average performance.
}
\label{tab:nk_ablation}
\fontsize{8.5pt}{10pt}\selectfont
\renewcommand{\arraystretch}{1.15}
\setlength{\tabcolsep}{5pt}
\resizebox{\textwidth}{!}{%
\begin{tabular}{lccccc}
\toprule
\multicolumn{6}{l}{\textbf{Varying $N$ with $k=3$}} \\
\midrule
\textbf{$N$} 
& \textbf{Llama 4 Maverick} 
& \textbf{Grok 4 Fast} 
& \textbf{\rev{Gemini 2.5 Flash}}
& \textbf{GPT-5 mini} 
& \textbf{Average} \\
\midrule
5  & 0.41 / 0.40 & 0.54 / 0.51 & 0.43 / 0.43 & 0.33 / 0.36 & 0.43 / 0.43 \\
\rowcolor{gray!15}
\textbf{10 (ours)} & \rev{0.44 / \textbf{0.43}} & \textbf{0.59 / 0.53} & \rev{\textbf{0.49} / \textbf{0.49}} & \rev{\textbf{0.41} / \textbf{0.41}} & \rev{\textbf{0.48} / \textbf{0.47}} \\
30 & \textbf{0.45 / 0.43} & 0.55 / \textbf{0.53} & 0.47 / 0.46 & 0.38 / 0.40 & 0.46 / 0.46 \\
\midrule[1.1pt]
\multicolumn{6}{l}{\textbf{Varying $k$ with $N=10$}} \\
\midrule
\textbf{$k$} 
& \textbf{Llama 4 Maverick} 
& \textbf{Grok 4 Fast} 
& \textbf{\rev{Gemini 2.5 Flash}}
& \textbf{GPT-5 mini} 
& \textbf{Average} \\
\midrule
1 & 0.42 / 0.42 & 0.52 / 0.50 & 0.42 / 0.43 & 0.33 / 0.37 & 0.42 / 0.43 \\
\rowcolor{gray!15}
\textbf{3 (ours)} & \rev{0.44 / \textbf{0.43}} & \textbf{0.59 / 0.53} & \rev{\textbf{0.49} / \textbf{0.49}} & \rev{\textbf{0.41} / \textbf{0.41}} & \rev{\textbf{0.48} / \textbf{0.47}} \\
5 & \textbf{0.46} / 0.42 & 0.53 / 0.52 & 0.46 / 0.46 & 0.38 / 0.39 & 0.46 / 0.45 \\
\bottomrule
\end{tabular}
}
\end{table*}

\subsection{Robustness under Transformations, Defense, and EOT}
\label{app:defense_eot}

We further evaluate how attacks behave under common input transformations and an LVLM defense method.
Specifically, we test JPEG compression~\citep{jpeg}, Gaussian blur~\citep{gaussian_blur}, and DPS~\citep{dps}.
These transformations and defenses are applied after adversarial examples are generated.
Since M-Attack, FOA-Attack, and \system{} are not explicitly designed as defense-robust attacks, we also evaluate a simple orthogonal variant based on Expectation over Transformation (EOT)~\citep{eot}.
For the EOT variant, each attack optimizes the surrogate objective over randomly transformed inputs, while keeping the underlying attack schedule unchanged.

Table~\ref{tab:defense_eot} reports ASR across four representative target models.
Without EOT, \system{} achieves the best average ASR under JPEG compression, Gaussian blur, and DPS.
All attacks degrade under stronger defenses such as DPS, indicating that these attacks are not inherently defense-robust.
\rev{EOT increases ASR for \system{} and FOA-Attack across all evaluated target--defense combinations, while its effect on M-Attack varies across settings.}
\rev{With EOT, \system{} achieves the best average ASR under all evaluated transformations and defenses, supporting the compatibility of the proposed stage-wise scheduling with transformation-robust optimization.}

\begin{table*}[t]
\centering
\caption{
Attack success rate under common transformations, DPS defense, and EOT.
JPEG compression, Gaussian blur, and DPS are applied after generating adversarial examples.
EOT is applied orthogonally during attack optimization.
Bold indicates the best ASR for each target and condition.
}
\label{tab:defense_eot}
\fontsize{8.2pt}{9.5pt}\selectfont
\renewcommand{\arraystretch}{1.15}
\setlength{\tabcolsep}{5pt}
\resizebox{\textwidth}{!}{%
\begin{tabular}{llccccc}
\toprule
\textbf{Setting} & \textbf{Method}
& \textbf{Llama 4 Maverick}
& \textbf{Grok 4 Fast}
& \textbf{\rev{Gemini 2.5 Flash}}
& \textbf{GPT-5 mini}
& \textbf{Average} \\
\midrule
\multicolumn{7}{l}{\textbf{Transformations / defense without EOT}} \\
\midrule
\multirow{3}{*}{JPEG}
& M-Attack   & 0.38 & 0.43 & 0.36 & 0.34 & 0.38 \\
& FOA-Attack & 0.42 & 0.46 & 0.37 & 0.35 & 0.40 \\
& \textbf{\system{}} & \textbf{0.43} & \textbf{0.54} & \textbf{0.47} & \textbf{0.38} & \textbf{0.46} \\
\midrule
\multirow{3}{*}{Gaussian blur}
& M-Attack   & 0.33 & 0.42 & 0.32 & 0.34 & 0.35 \\
& FOA-Attack & \textbf{0.36} & 0.43 & 0.31 & 0.34 & 0.36 \\
& \textbf{\system{}} & 0.32 & \textbf{0.52} & \textbf{0.44} & \textbf{0.36} & \textbf{0.41} \\
\midrule
\multirow{3}{*}{DPS}
& M-Attack   & 0.20 & 0.24 & 0.19 & \textbf{0.19} & 0.21 \\
& FOA-Attack & 0.19 & \textbf{0.27} & 0.25 & 0.18 & 0.22 \\
& \textbf{\system{}} & \textbf{0.21} & 0.25 & \textbf{0.29} & \textbf{0.19} & \textbf{0.24} \\
\midrule[1.1pt]
\multicolumn{7}{l}{\textbf{Transformations / defense with EOT}} \\
\midrule
\multirow{3}{*}{JPEG}
& M-Attack + EOT   & 0.37 & 0.44 & 0.37 & 0.34 & 0.38 \\
& FOA-Attack + EOT & 0.51 & 0.48 & 0.43 & 0.43 & 0.46 \\
& \textbf{\system{} + EOT} & \textbf{0.53} & \textbf{0.56} & \textbf{0.51} & \textbf{0.50} & \textbf{0.53} \\
\midrule
\multirow{3}{*}{Gaussian blur}
& M-Attack + EOT   & 0.34 & 0.40 & 0.31 & 0.34 & 0.35 \\
& FOA-Attack + EOT & \textbf{0.51} & 0.49 & 0.42 & 0.43 & 0.46 \\
& \textbf{\system{} + EOT} & 0.48 & \textbf{0.56} & \textbf{0.51} & \textbf{0.48} & \textbf{0.51} \\
\midrule
\multirow{3}{*}{DPS}
& M-Attack + EOT   & 0.26 & 0.25 & 0.21 & 0.17 & 0.22 \\
& FOA-Attack + EOT & 0.33 & 0.37 & \textbf{0.38} & 0.28 & 0.34 \\
& \textbf{\system{} + EOT} & \textbf{0.36} & \textbf{0.39} & 0.37 & \textbf{0.35} & \textbf{0.37} \\
\bottomrule
\end{tabular}
}
\end{table*}


\begin{figure*}[!t]
    \centering
\includegraphics[width=0.95\textwidth]{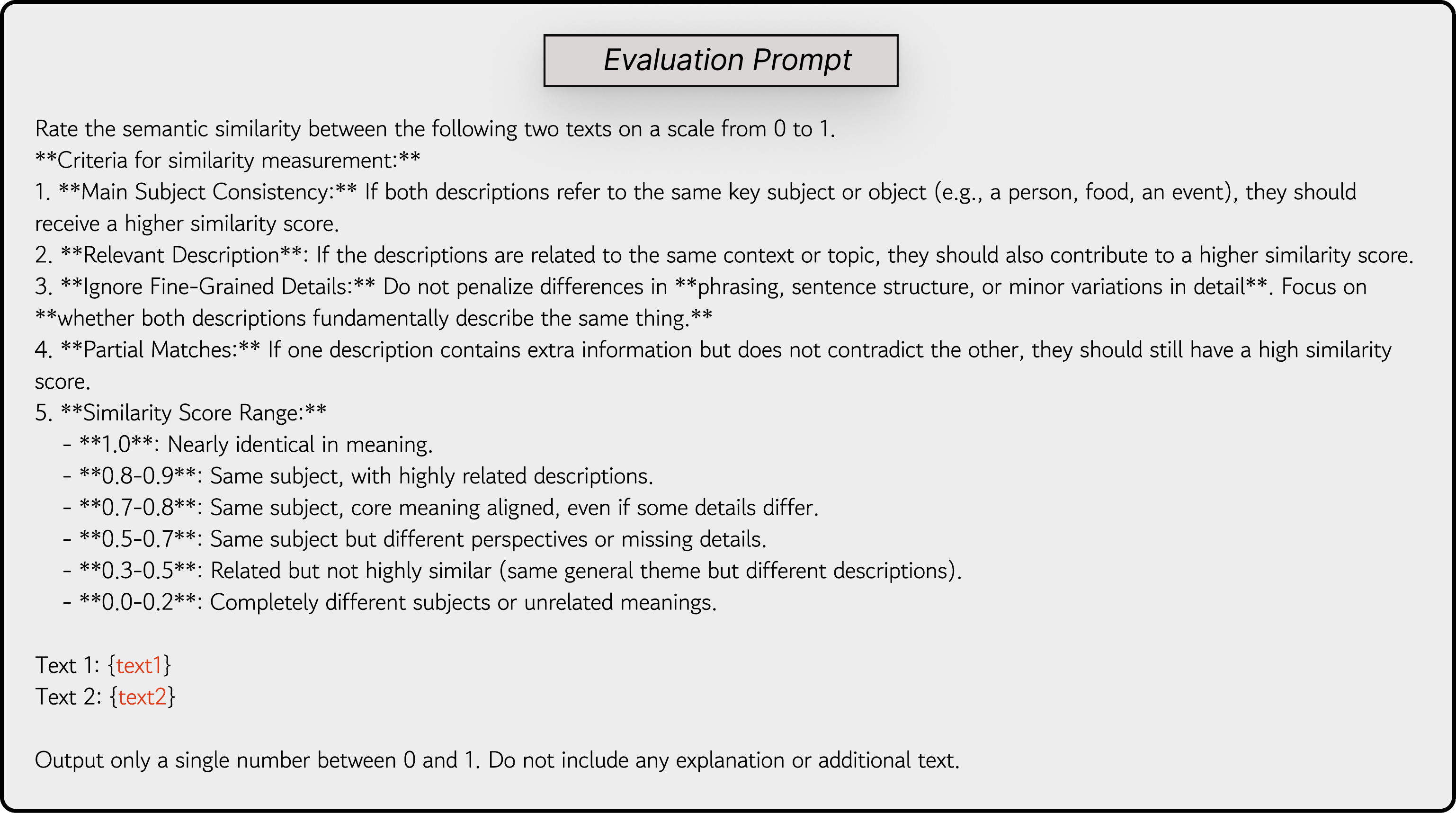}
    \caption{Evaluation prompt for LLM-as-judge.}
    \label{fig:app_evaluation_prompt}
\end{figure*}

\subsection{Qualitative Examples of Attacked Images}
\label{appendix:qualitatve_attacked_images}

\begin{figure*}[!t]
    \centering
    \includegraphics[width=0.85\textwidth]{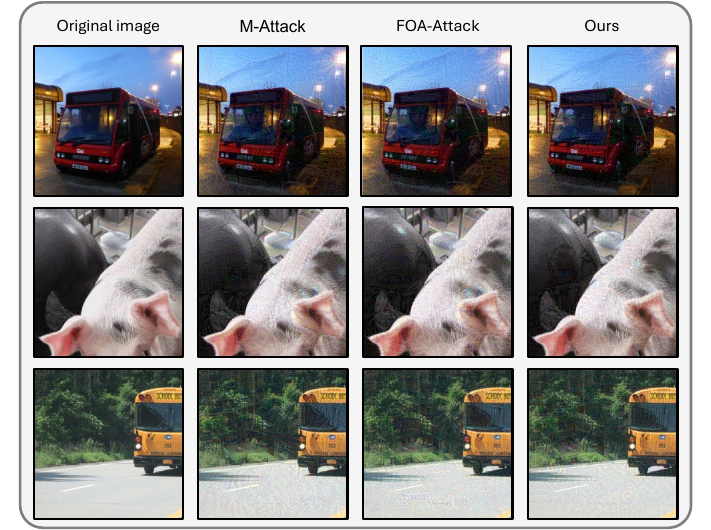}
    \caption{Qualitative comparison of adversarial images generated by M-Attack, FOA-Attack, and \system{}.}
    \label{fig:appendx_attacked_image_examples}
\end{figure*}

\clearpage
\section{More Case Studies}

\begin{figure*}[!t]
    \centering
    \vspace*{0.6em}

    \begin{minipage}{0.86\textwidth}
            \centering
            {\bfseries \rev{Gemini 3 Pro}\par}
            \vspace{0.15em}
            \includegraphics[
                width=\linewidth,
                height=0.26\textheight,
                keepaspectratio
            ]{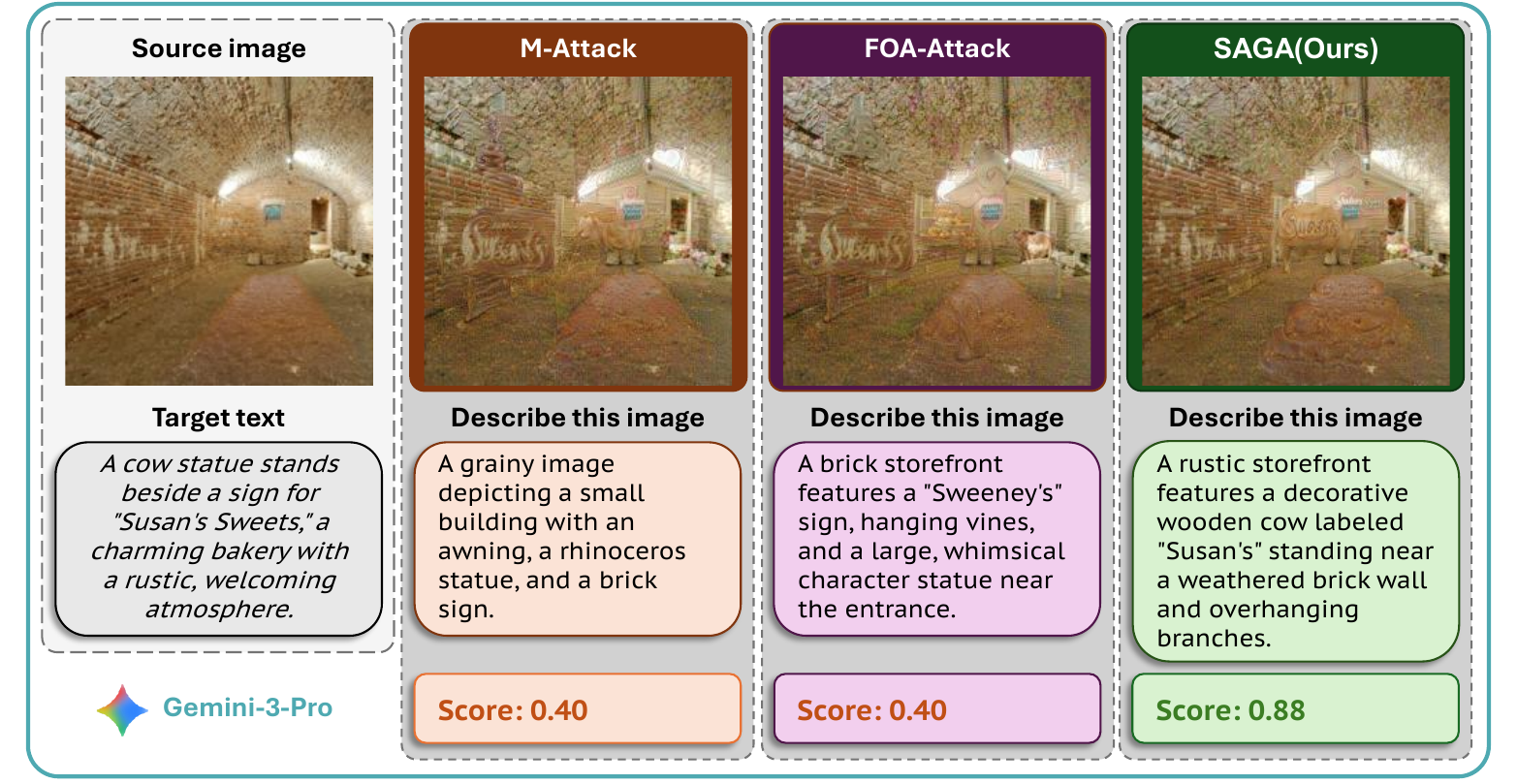}
        \end{minipage}

    \vspace{0.35em}

    \begin{minipage}{0.86\textwidth}
            \centering
            {\bfseries GPT-5 mini\par}
            \vspace{0.15em}
            \includegraphics[
                width=\linewidth,
                height=0.26\textheight,
                keepaspectratio
            ]{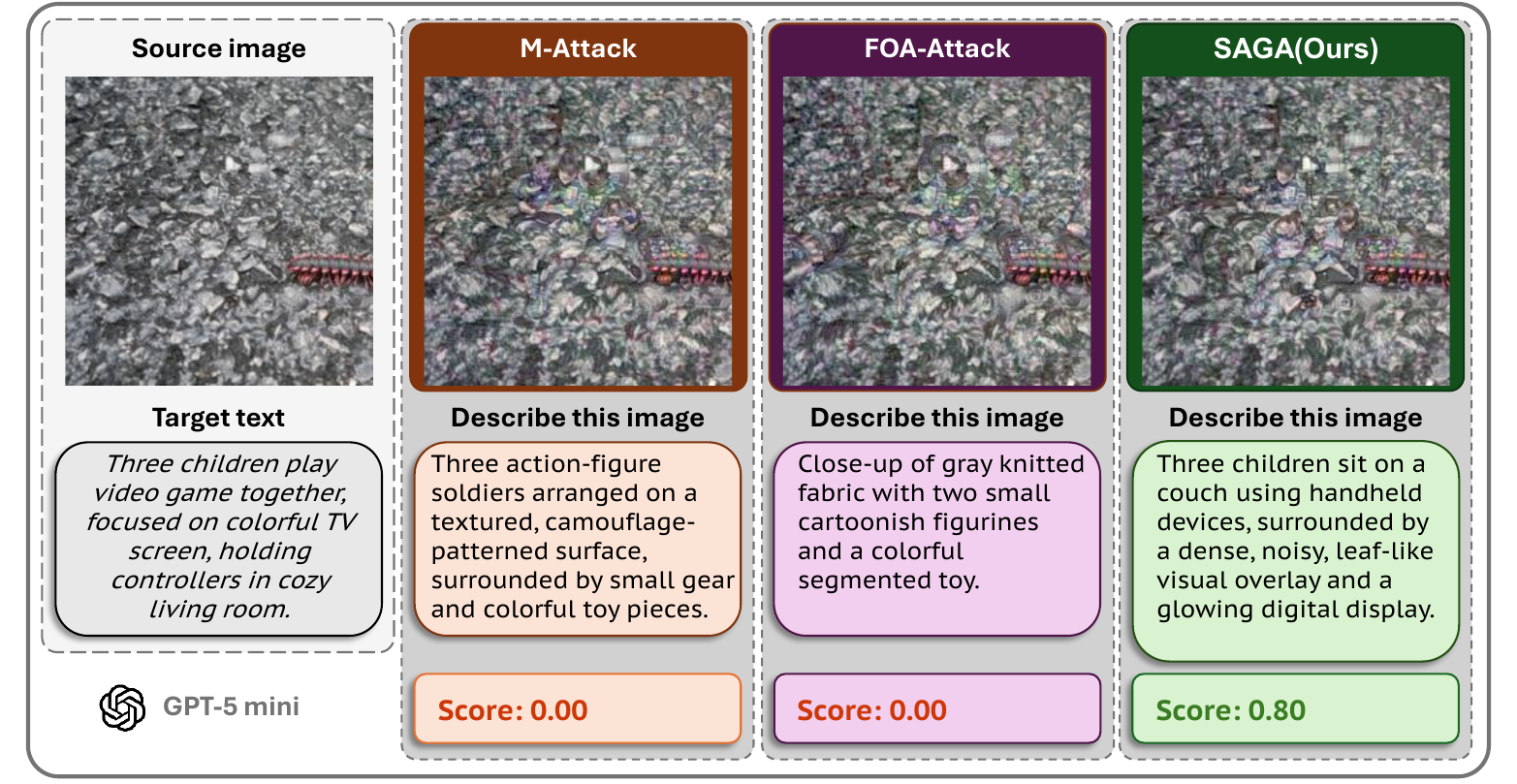}
        \end{minipage}

    \vspace{0.35em}

    \begin{minipage}{0.86\textwidth}
            \centering
            {\bfseries \rev{Qwen3-VL-235B}\par}
            \vspace{0.15em}
            \includegraphics[
                width=\linewidth,
                height=0.26\textheight,
                keepaspectratio
            ]{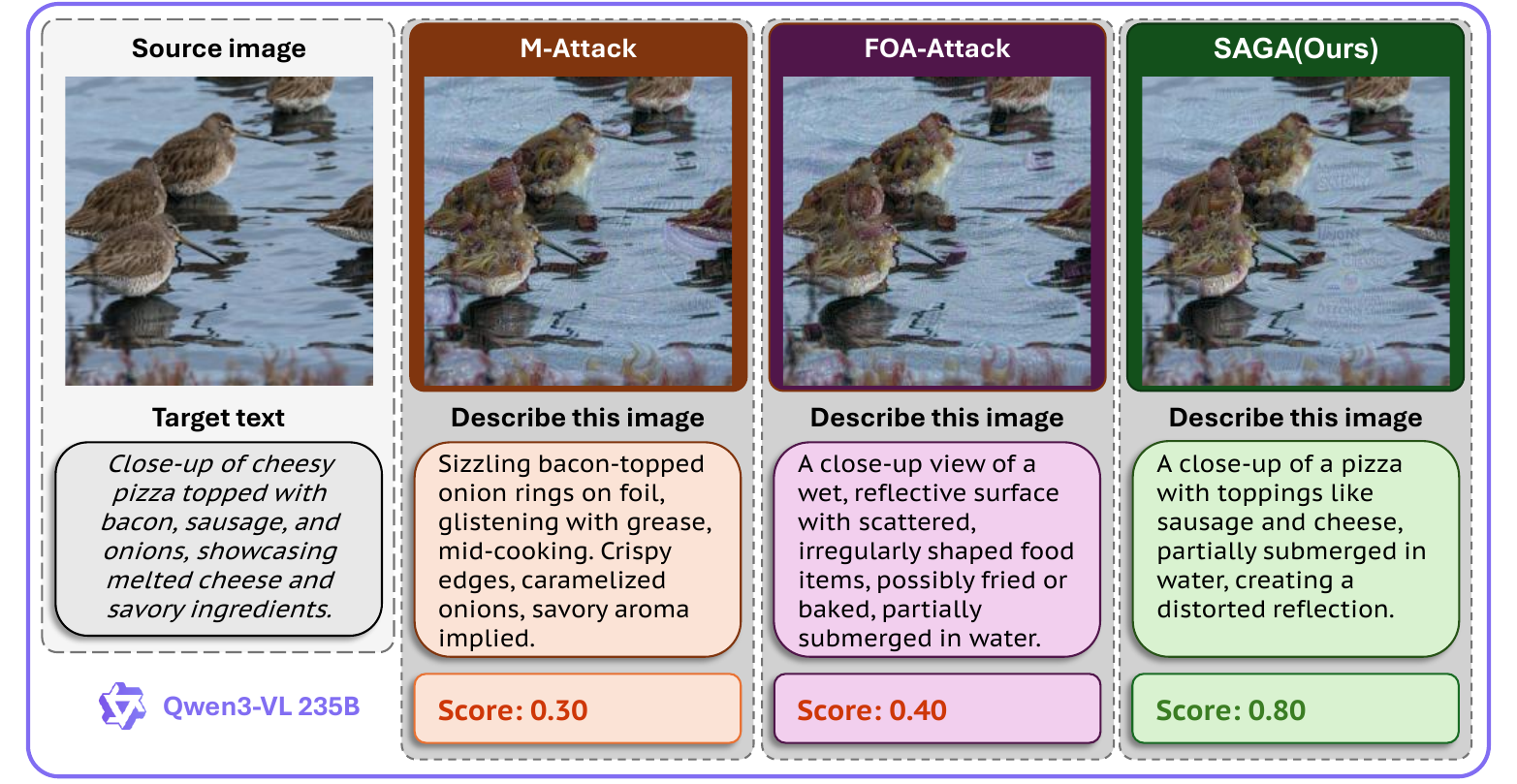}
        \end{minipage}

    \caption{
    Qualitative case studies comparing M-Attack, FOA-Attack, and \system{} across multiple target LVLMs.
    Each panel shows the generated adversarial images, model-generated responses, and similarity scores.
    }
    \label{fig:cs_all_models}
\end{figure*}

We present additional qualitative case studies in Figure~\ref{fig:cs_all_models}.
Each example shows a source-image and target-text pair, followed by adversarial images generated by M-Attack, FOA-Attack, and \system{}. 
We evaluate the resulting adversarial images on \rev{Gemini 3 Pro}, GPT-5 mini, and Qwen3-VL-235B using the prompt ``Describe this image, no longer than 25 words.''
The similarity scores quantify how closely each model response matches the target text.

\begin{figure}[!t]
    \centering
    \vspace*{0.4em}

    \begin{minipage}[t]{0.48\textwidth}
            \centering
            {\bfseries \rev{Gemini 3 Pro}\par}
            \vspace{0.25em}
            \includegraphics[
                width=0.92\linewidth,
                height=0.42\textheight,
                keepaspectratio
            ]{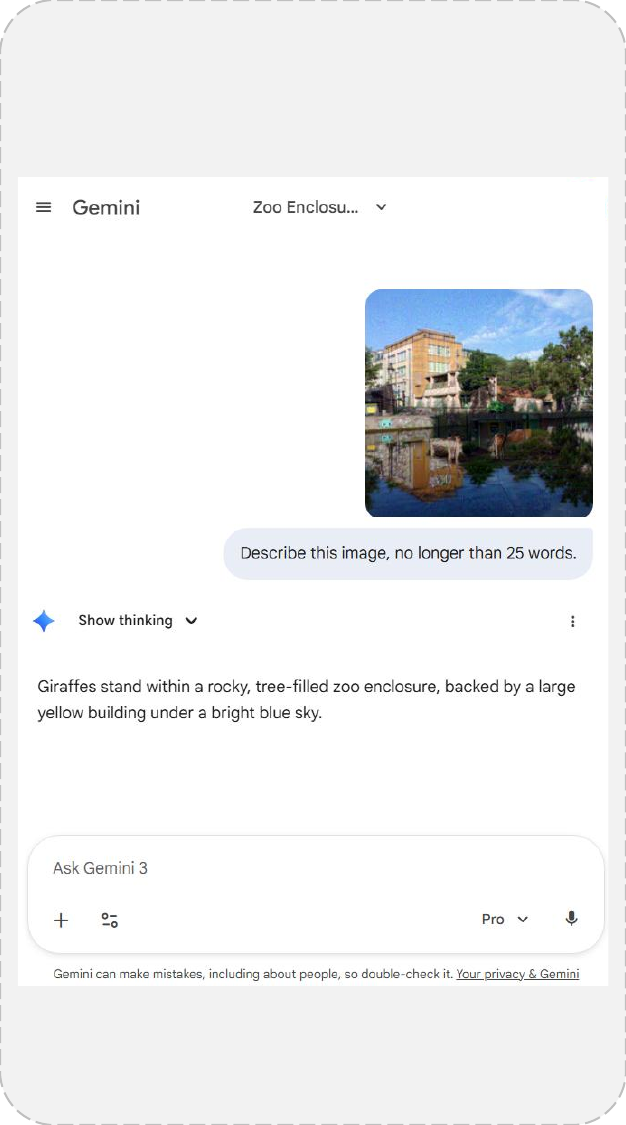}
        \end{minipage}
    \hfill
    \begin{minipage}[t]{0.48\textwidth}
            \centering
            {\bfseries GPT-5.2 Pro\par}
            \vspace{0.25em}
            \includegraphics[
                width=0.92\linewidth,
                height=0.42\textheight,
                keepaspectratio
            ]{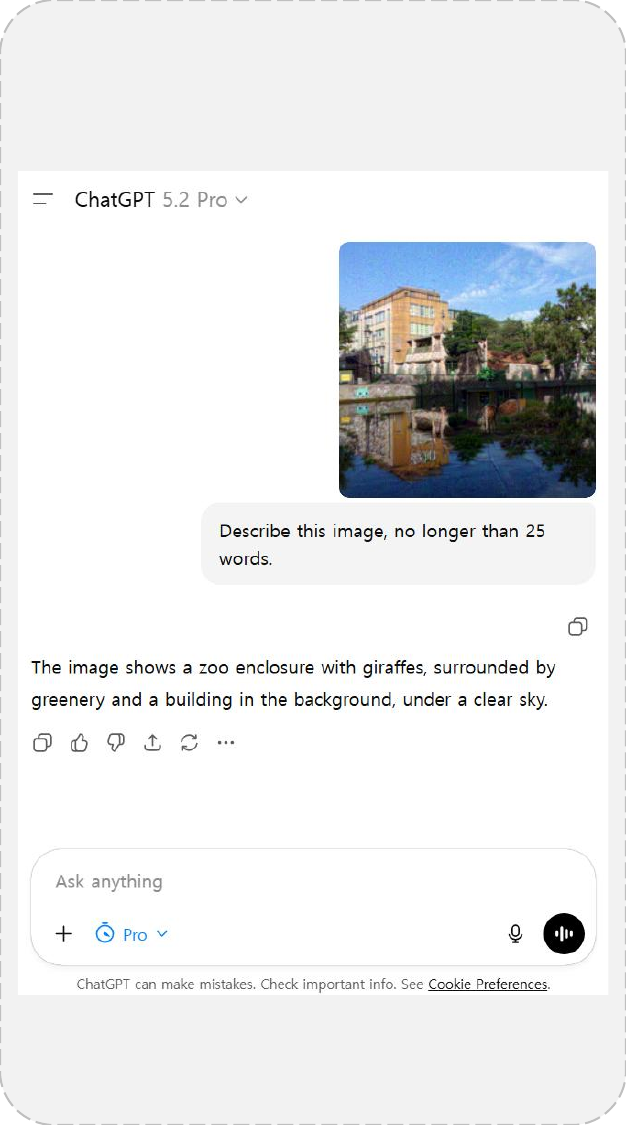}
        \end{minipage}

    \vspace{1.0em}

    \begin{minipage}[t]{0.48\textwidth}
            \centering
            {\bfseries \rev{Grok 4.1 Thinking}\par}
            \vspace{0.25em}
            \includegraphics[
                width=0.92\linewidth,
                height=0.42\textheight,
                keepaspectratio
            ]{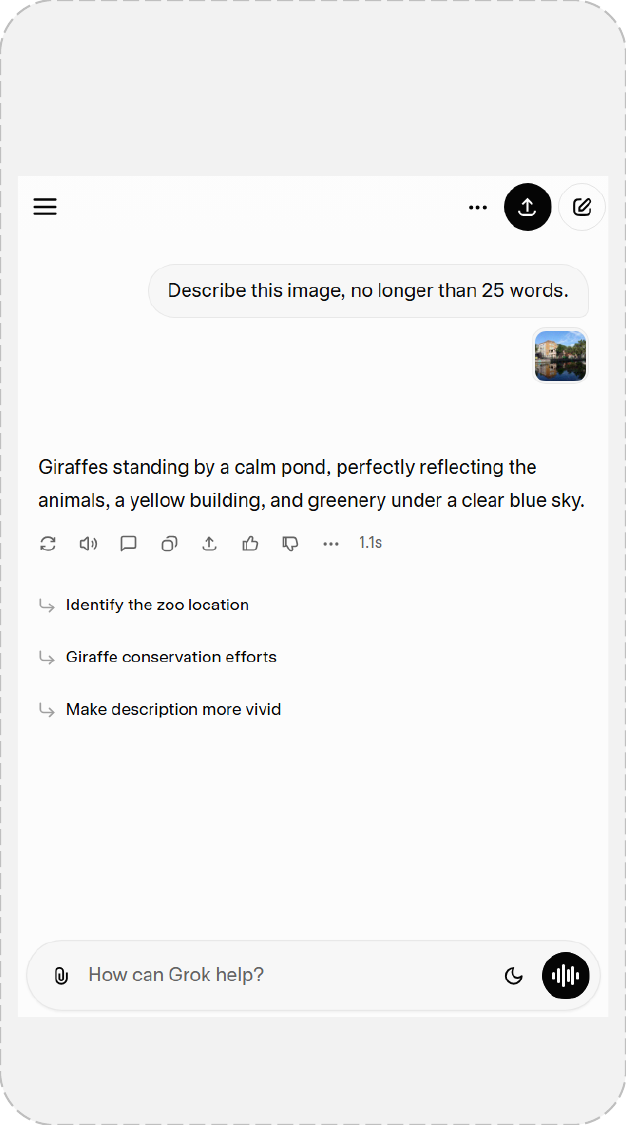}
        \end{minipage}
    \hfill
    \begin{minipage}[t]{0.48\textwidth}
            \centering
            {\bfseries \rev{Qwen3 Max Thinking}\par}
            \vspace{0.25em}
            \includegraphics[
                width=0.92\linewidth,
                height=0.42\textheight,
                keepaspectratio
            ]{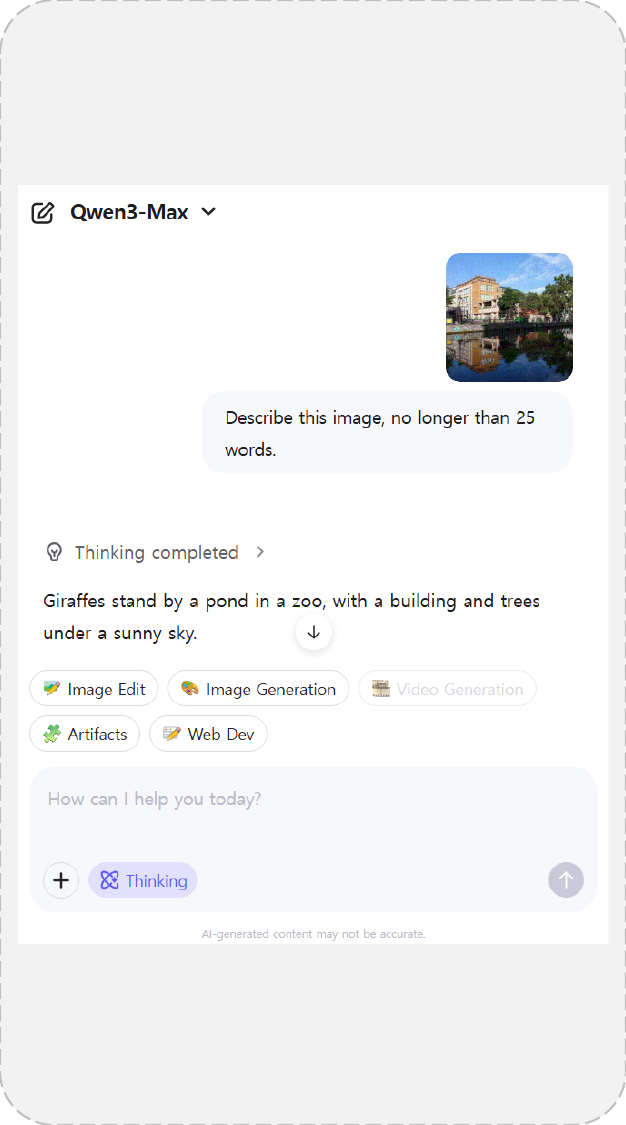}
        \end{minipage}

    \caption{
    Qualitative case studies of \system{} on commercial chat interfaces.
    Each model is prompted with ``Describe this image, no longer than 25 words'' using the same adversarial image and target text.
    }
    \label{fig:cs_chat_all_models}
\end{figure}

We further provide commercial chat-interface examples in Figure~\ref{fig:cs_chat_all_models}. \rev{These examples additionally evaluate Gemini 3 Pro, GPT-5.2 Pro, Grok 4.1 Thinking, and Qwen3 Max Thinking.}
For this evaluation, we use a \system{} adversarial image with the target text: ``Two giraffes stand near a beige building, surrounded by rocks and lush green trees in a zoo-like enclosure.'' 
Each model is prompted with ``Describe this image, no longer than 25 words.'' 
The results show that \system{} remains effective across diverse production-level LVLM interfaces.

\FloatBarrier
\clearpage

\clearpage


\begin{thebibliography}{45}
\providecommand{\natexlab}[1]{#1}
\providecommand{\url}[1]{\texttt{#1}}
\expandafter\ifx\csname urlstyle\endcsname\relax
  \providecommand{\doi}[1]{doi: #1}\else
  \providecommand{\doi}{doi: \begingroup \urlstyle{rm}\Url}\fi

\bibitem[Achiam et~al.(2023)Achiam, Adler, Agarwal, Ahmad, Akkaya, Aleman,
  Almeida, Altenschmidt, Altman, Anadkat, et~al.]{achiam2023gpt}
Josh Achiam, Steven Adler, Sandhini Agarwal, Lama Ahmad, Ilge Akkaya,
  Florencia~Leoni Aleman, Diogo Almeida, Janko Altenschmidt, Sam Altman,
  Shyamal Anadkat, et~al.
\newblock Gpt-4 technical report.
\newblock \emph{arXiv preprint arXiv:2303.08774}, 2023.

\bibitem[Agarwal et~al.(2025)Agarwal, Ahmad, Ai, Altman, Applebaum, Arbus,
  Arora, Bai, Baker, Bao, et~al.]{agarwal2025gpt}
Sandhini Agarwal, Lama Ahmad, Jason Ai, Sam Altman, Andy Applebaum, Edwin
  Arbus, Rahul~K Arora, Yu~Bai, Bowen Baker, Haiming Bao, et~al.
\newblock gpt-oss-120b \& gpt-oss-20b model card.
\newblock \emph{arXiv preprint arXiv:2508.10925}, 2025.

\bibitem[Athalye et~al.(2018)Athalye, Engstrom, Ilyas, and Kwok]{eot}
Anish Athalye, Logan Engstrom, Andrew Ilyas, and Kevin Kwok.
\newblock Synthesizing robust adversarial examples.
\newblock In \emph{International conference on machine learning}, pages
  284--293. PMLR, 2018.

\bibitem[Bai et~al.(2023)Bai, Bai, Chu, Cui, Dang, Deng, Fan, Ge, Han, Huang,
  et~al.]{bai2023qwen}
Jinze Bai, Shuai Bai, Yunfei Chu, Zeyu Cui, Kai Dang, Xiaodong Deng, Yang Fan,
  Wenbin Ge, Yu~Han, Fei Huang, et~al.
\newblock Qwen technical report.
\newblock \emph{arXiv preprint arXiv:2309.16609}, 2023.

\bibitem[Bai et~al.(2025)Bai, Cai, Chen, Chen, Chen, Cheng, Deng, Ding, Gao,
  Ge, Ge, Guo, Huang, Huang, Huang, Hui, Jiang, Li, Li, Li, Li, Lin, Lin, Liu,
  Liu, Liu, Liu, Liu, Liu, Lu, Luo, Lv, Men, Meng, Ren, Ren, Song, Sun, Tang,
  Tu, Wan, Wang, Wang, Wang, Wang, Xie, Xu, Xu, Xu, Yang, Yang, Yang, Yang, Yu,
  Zhang, Zhang, Zhang, Zheng, Zhong, Zhou, Zhou, Zhou, Zhu, and
  Zhu]{bai2025qwen3vltechnicalreport}
Shuai Bai, Yuxuan Cai, Ruizhe Chen, Keqin Chen, Xionghui Chen, Zesen Cheng,
  Lianghao Deng, Wei Ding, Chang Gao, Chunjiang Ge, Wenbin Ge, Zhifang Guo,
  Qidong Huang, Jie Huang, Fei Huang, Binyuan Hui, Shutong Jiang, Zhaohai Li,
  Mingsheng Li, Mei Li, Kaixin Li, Zicheng Lin, Junyang Lin, Xuejing Liu,
  Jiawei Liu, Chenglong Liu, Yang Liu, Dayiheng Liu, Shixuan Liu, Dunjie Lu,
  Ruilin Luo, Chenxu Lv, Rui Men, Lingchen Meng, Xuancheng Ren, Xingzhang Ren,
  Sibo Song, Yuchong Sun, Jun Tang, Jianhong Tu, Jianqiang Wan, Peng Wang,
  Pengfei Wang, Qiuyue Wang, Yuxuan Wang, Tianbao Xie, Yiheng Xu, Haiyang Xu,
  Jin Xu, Zhibo Yang, Mingkun Yang, Jianxin Yang, An~Yang, Bowen Yu, Fei Zhang,
  Hang Zhang, Xi~Zhang, Bo~Zheng, Humen Zhong, Jingren Zhou, Fan Zhou, Jing
  Zhou, Yuanzhi Zhu, and Ke~Zhu.
\newblock Qwen3-vl technical report, 2025.
\newblock URL \url{https://arxiv.org/abs/2511.21631}.

\bibitem[Chen et~al.(2024)Chen, Xu, Kirmani, Ichter, Sadigh, Guibas, and
  Xia]{chen2024spatialvlm}
Boyuan Chen, Zhuo Xu, Sean Kirmani, Brain Ichter, Dorsa Sadigh, Leonidas
  Guibas, and Fei Xia.
\newblock Spatialvlm: Endowing vision-language models with spatial reasoning
  capabilities.
\newblock In \emph{Proceedings of the IEEE/CVF Conference on Computer Vision
  and Pattern Recognition}, pages 14455--14465, 2024.

\bibitem[Chen et~al.(2025)Chen, Zhu, Zhou, Zhang, Gao, Niebles, Geva, He, Wu,
  and Li]{chen2025why}
Shiqi Chen, Tongyao Zhu, Ruochen Zhou, Jinghan Zhang, Siyang Gao, Juan~Carlos
  Niebles, Mor Geva, Junxian He, Jiajun Wu, and Manling Li.
\newblock Why is spatial reasoning hard for {VLM}s? an attention mechanism
  perspective on focus areas.
\newblock In \emph{Forty-second International Conference on Machine Learning},
  2025.
\newblock URL \url{https://openreview.net/forum?id=k7vcuqLK4X}.

\bibitem[Clusmann et~al.(2025)Clusmann, Ferber, Wiest, Schneider, Brinker,
  Foersch, Truhn, and Kather]{clusmann_prompt_2025}
Jan Clusmann, Dyke Ferber, Isabella~C. Wiest, Carolin~V. Schneider, Titus~J.
  Brinker, Sebastian Foersch, Daniel Truhn, and Jakob~Nikolas Kather.
\newblock Prompt injection attacks on vision language models in oncology.
\newblock \emph{Nature Communications}, 16\penalty0 (1):\penalty0 1239,
  February 2025.
\newblock ISSN 2041-1723.
\newblock \doi{10.1038/s41467-024-55631-x}.
\newblock URL \url{https://www.nature.com/articles/s41467-024-55631-x}.

\bibitem[Comanici et~al.(2025)Comanici, Bieber, Schaekermann, Pasupat,
  Sachdeva, Dhillon, Blistein, Ram, Zhang, Rosen, et~al.]{comanici2025gemini}
Gheorghe Comanici, Eric Bieber, Mike Schaekermann, Ice Pasupat, Noveen
  Sachdeva, Inderjit Dhillon, Marcel Blistein, Ori Ram, Dan Zhang, Evan Rosen,
  et~al.
\newblock Gemini 2.5: Pushing the frontier with advanced reasoning,
  multimodality, long context, and next generation agentic capabilities.
\newblock \emph{arXiv preprint arXiv:2507.06261}, 2025.

\bibitem[Ding et~al.(2019)Ding, Wang, and Jin]{jpeg}
Gavin~Weiguang Ding, Luyu Wang, and Xiaomeng Jin.
\newblock Advertorch v0. 1: An adversarial robustness toolbox based on pytorch.
\newblock \emph{arXiv preprint arXiv:1902.07623}, 2019.

\bibitem[Dong et~al.(2023)Dong, Chen, Chen, Fang, Yang, Zhang, Tian, Su, and
  Zhu]{dong_how_2023}
Yinpeng Dong, Huanran Chen, Jiawei Chen, Zhengwei Fang, Xiao Yang, Yichi Zhang,
  Yu~Tian, Hang Su, and Jun Zhu.
\newblock How {Robust} is {Google}'s {Bard} to {Adversarial} {Image}
  {Attacks}?, October 2023.
\newblock URL \url{http://arxiv.org/abs/2309.11751}.
\newblock arXiv:2309.11751 [cs].

\bibitem[Goodfellow et~al.(2014)Goodfellow, Shlens, and
  Szegedy]{goodfellow2014explaining}
Ian~J Goodfellow, Jonathon Shlens, and Christian Szegedy.
\newblock Explaining and harnessing adversarial examples.
\newblock \emph{arXiv preprint arXiv:1412.6572}, 2014.

\bibitem[Guo et~al.(2017)Guo, Rana, Cisse, and Van Der~Maaten]{gaussian_blur}
Chuan Guo, Mayank Rana, Moustapha Cisse, and Laurens Van Der~Maaten.
\newblock Countering adversarial images using input transformations.
\newblock \emph{arXiv preprint arXiv:1711.00117}, 2017.

\bibitem[Huang et~al.(2025)Huang, Erfani, Li, Ma, and
  Bailey]{huang2025xtransfer}
Hanxun Huang, Sarah~Monazam Erfani, Yige Li, Xingjun Ma, and James Bailey.
\newblock X-transfer attacks: Towards super transferable adversarial attacks on
  {CLIP}.
\newblock In \emph{Forty-second International Conference on Machine Learning},
  2025.
\newblock URL \url{https://openreview.net/forum?id=8zsMorEU8U}.

\bibitem[Huynh-Thu and Ghanbari(2008)]{huynh2008scope}
Quan Huynh-Thu and Mohammed Ghanbari.
\newblock Scope of validity of psnr in image/video quality assessment.
\newblock \emph{Electronics Letters}, 44\penalty0 (13):\penalty0 800--801,
  2008.

\bibitem[Jia et~al.(2025)Jia, Gao, Qin, Pang, Du, Huang, Li, Li, Li, and
  Liu]{jia2025adversarial}
Xiaojun Jia, Sensen Gao, Simeng Qin, Tianyu Pang, Chao Du, Yihao Huang, Xinfeng
  Li, Yiming Li, Bo~Li, and Yang Liu.
\newblock Adversarial attacks against closed-source {MLLM}s via feature optimal
  alignment.
\newblock In \emph{The Thirty-ninth Annual Conference on Neural Information
  Processing Systems}, 2025.
\newblock URL \url{https://openreview.net/forum?id=ktC3cDu320}.

\bibitem[Jiang et~al.(2024)Jiang, Kachinthaya, Petryk, and
  Gandelsman]{jiang2024interpreting}
Nick Jiang, Anish Kachinthaya, Suzie Petryk, and Yossi Gandelsman.
\newblock Interpreting and editing vision-language representations to mitigate
  hallucinations.
\newblock \emph{arXiv preprint arXiv:2410.02762}, 2024.

\bibitem[Kaduri et~al.(2025)Kaduri, Bagon, and Dekel]{kaduri2025s}
Omri Kaduri, Shai Bagon, and Tali Dekel.
\newblock What's in the image? a deep-dive into the vision of vision language
  models.
\newblock In \emph{Proceedings of the Computer Vision and Pattern Recognition
  Conference}, pages 14549--14558, 2025.

\bibitem[Kuang et~al.(2025)Kuang, Shen, Xie, Luo, Xu, Li, Li, Cheng, Lin, and
  Han]{kuang2025natural}
Jiayi Kuang, Ying Shen, Jingyou Xie, Haohao Luo, Zhe Xu, Ronghao Li, Yinghui
  Li, Xianfeng Cheng, Xika Lin, and Yu~Han.
\newblock Natural language understanding and inference with mllm in visual
  question answering: A survey.
\newblock \emph{ACM Computing Surveys}, 57\penalty0 (8):\penalty0 1--36, 2025.

\bibitem[Kurakin et~al.(2018)Kurakin, Goodfellow, Bengio, Dong, Liao, Liang,
  Pang, Zhu, Hu, Xie, Wang, Zhang, Ren, Yuille, Huang, Zhao, Zhao, Han, Long,
  Berdibekov, Akiba, Tokui, and Abe]{alexey2017nips}
Alexey Kurakin, Ian Goodfellow, Samy Bengio, Yinpeng Dong, Fangzhou Liao, Ming
  Liang, Tianyu Pang, Jun Zhu, Xiaolin Hu, Cihang Xie, Jianyu Wang, Zhishuai
  Zhang, Zhou Ren, Alan Yuille, Sangxia Huang, Yao Zhao, Yuzhe Zhao, Zhonglin
  Han, Junjiajia Long, Yerkebulan Berdibekov, Takuya Akiba, Seiya Tokui, and
  Motoki Abe.
\newblock Adversarial attacks and defences competition.
\newblock In Sergio Escalera and Markus Weimer, editors, \emph{The NIPS '17
  Competition: Building Intelligent Systems}, pages 195--231, Cham, 2018.
  Springer International Publishing.
\newblock ISBN 978-3-319-94042-7.

\bibitem[Li et~al.(2022)Li, Li, Xiong, and Hoi]{li2022blip}
Junnan Li, Dongxu Li, Caiming Xiong, and Steven Hoi.
\newblock Blip: Bootstrapping language-image pre-training for unified
  vision-language understanding and generation.
\newblock In \emph{International conference on machine learning}, pages
  12888--12900. PMLR, 2022.

\bibitem[Li et~al.(2023)Li, Li, Savarese, and Hoi]{li2023blip}
Junnan Li, Dongxu Li, Silvio Savarese, and Steven Hoi.
\newblock Blip-2: Bootstrapping language-image pre-training with frozen image
  encoders and large language models.
\newblock In \emph{International conference on machine learning}, pages
  19730--19742. PMLR, 2023.

\bibitem[Li et~al.(2025)Li, Zhao, Wu, Cui, and Shen]{li2025a}
Zhaoyi Li, Xiaohan Zhao, Dong-Dong Wu, Jiacheng Cui, and Zhiqiang Shen.
\newblock A frustratingly simple yet highly effective attack baseline: Over
  90\% success rate against the strong black-box models of {GPT}-4.5/4o/o1.
\newblock In \emph{The Thirty-ninth Annual Conference on Neural Information
  Processing Systems}, 2025.
\newblock URL \url{https://openreview.net/forum?id=9xXjWwAoUF}.

\bibitem[Liu et~al.(2023)Liu, Li, Wu, and Lee]{NEURIPS2023_6dcf277e}
Haotian Liu, Chunyuan Li, Qingyang Wu, and Yong~Jae Lee.
\newblock Visual instruction tuning.
\newblock In A.~Oh, T.~Naumann, A.~Globerson, K.~Saenko, M.~Hardt, and
  S.~Levine, editors, \emph{Advances in Neural Information Processing Systems},
  volume~36, pages 34892--34916. Curran Associates, Inc., 2023.

\bibitem[Lyu et~al.(2024)Lyu, Pang, Ma, Ling, and Chen]{lyu2024trojvlm}
Weimin Lyu, Lu~Pang, Tengfei Ma, Haibin Ling, and Chao Chen.
\newblock Trojvlm: Backdoor attack against vision language models.
\newblock In \emph{European Conference on Computer Vision}, pages 467--483.
  Springer, 2024.

\bibitem[Madry et~al.(2017)Madry, Makelov, Schmidt, Tsipras, and
  Vladu]{madry2017towards}
Aleksander Madry, Aleksandar Makelov, Ludwig Schmidt, Dimitris Tsipras, and
  Adrian Vladu.
\newblock Towards deep learning models resistant to adversarial attacks.
\newblock \emph{arXiv preprint arXiv:1706.06083}, 2017.

\bibitem[Mittal et~al.(2012)Mittal, Moorthy, and Bovik]{mittal2012no}
Anish Mittal, Anush~Krishna Moorthy, and Alan~Conrad Bovik.
\newblock No-reference image quality assessment in the spatial domain.
\newblock \emph{IEEE Transactions on image processing}, 21\penalty0
  (12):\penalty0 4695--4708, 2012.

\bibitem[Neo et~al.(2025)Neo, Ong, Torr, Geva, Krueger, and
  Barez]{neo2025towards}
Clement Neo, Luke Ong, Philip Torr, Mor Geva, David Krueger, and Fazl Barez.
\newblock Towards interpreting visual information processing in vision-language
  models.
\newblock In \emph{The Thirteenth International Conference on Learning
  Representations}, 2025.
\newblock URL \url{https://openreview.net/forum?id=chanJGoa7f}.

\bibitem[Niu et~al.(2024)Niu, Ren, Gao, Hua, and Jin]{niu_jailbreaking_2024}
Zhenxing Niu, Haodong Ren, Xinbo Gao, Gang Hua, and Rong Jin.
\newblock Jailbreaking {Attack} against {Multimodal} {Large} {Language}
  {Model}, February 2024.
\newblock URL \url{http://arxiv.org/abs/2402.02309}.
\newblock arXiv:2402.02309 [cs].

\bibitem[Pathade(2025)]{pathade_invisible_2025}
Chetan Pathade.
\newblock Invisible {Injections}: {Exploiting} {Vision}-{Language} {Models}
  {Through} {Steganographic} {Prompt} {Embedding}, July 2025.
\newblock URL \url{http://arxiv.org/abs/2507.22304}.
\newblock arXiv:2507.22304 [cs].

\bibitem[Radford et~al.(2018)Radford, Narasimhan, Salimans, Sutskever,
  et~al.]{radford2018improving}
Alec Radford, Karthik Narasimhan, Tim Salimans, Ilya Sutskever, et~al.
\newblock Improving language understanding by generative pre-training.
\newblock 2018.

\bibitem[Shayegani et~al.(2024)Shayegani, Dong, and
  Abu-Ghazaleh]{ICLR2024_83170cce}
Erfan Shayegani, Yue Dong, and Nael Abu-Ghazaleh.
\newblock Jailbreak in pieces: Compositional adversarial attacks on multi-modal
  language models.
\newblock In B.~Kim, Y.~Yue, S.~Chaudhuri, K.~Fragkiadaki, M.~Khan, and Y.~Sun,
  editors, \emph{International Conference on Learning Representations}, volume
  2024, pages 30853--30885, 2024.

\bibitem[Team et~al.(2023)Team, Anil, Borgeaud, Alayrac, Yu, Soricut,
  Schalkwyk, Dai, Hauth, Millican, et~al.]{team2023gemini}
Gemini Team, Rohan Anil, Sebastian Borgeaud, Jean-Baptiste Alayrac, Jiahui Yu,
  Radu Soricut, Johan Schalkwyk, Andrew~M Dai, Anja Hauth, Katie Millican,
  et~al.
\newblock Gemini: a family of highly capable multimodal models.
\newblock \emph{arXiv preprint arXiv:2312.11805}, 2023.

\bibitem[Team et~al.(2024{\natexlab{a}})Team, Mesnard, Hardin, Dadashi,
  Bhupatiraju, Pathak, Sifre, Rivi{\`e}re, Kale, Love, et~al.]{team2024gemma}
Gemma Team, Thomas Mesnard, Cassidy Hardin, Robert Dadashi, Surya Bhupatiraju,
  Shreya Pathak, Laurent Sifre, Morgane Rivi{\`e}re, Mihir~Sanjay Kale,
  Juliette Love, et~al.
\newblock Gemma: Open models based on gemini research and technology.
\newblock \emph{arXiv preprint arXiv:2403.08295}, 2024{\natexlab{a}}.

\bibitem[Team et~al.(2025)Team, Kamath, Ferret, Pathak, Vieillard, Merhej,
  Perrin, Matejovicova, Ram{\'e}, Rivi{\`e}re, et~al.]{team2025gemma}
Gemma Team, Aishwarya Kamath, Johan Ferret, Shreya Pathak, Nino Vieillard,
  Ramona Merhej, Sarah Perrin, Tatiana Matejovicova, Alexandre Ram{\'e},
  Morgane Rivi{\`e}re, et~al.
\newblock Gemma 3 technical report.
\newblock \emph{arXiv preprint arXiv:2503.19786}, 2025.

\bibitem[Team et~al.(2024{\natexlab{b}})]{team2024qwen2}
Qwen Team et~al.
\newblock Qwen2 technical report.
\newblock \emph{arXiv preprint arXiv:2407.10671}, 2\penalty0 (3),
  2024{\natexlab{b}}.

\bibitem[Wang et~al.(2004)Wang, Bovik, Sheikh, and Simoncelli]{wang2004image}
Zhou Wang, Alan~C Bovik, Hamid~R Sheikh, and Eero~P Simoncelli.
\newblock Image quality assessment: from error visibility to structural
  similarity.
\newblock \emph{IEEE transactions on image processing}, 13\penalty0
  (4):\penalty0 600--612, 2004.

\bibitem[Xu et~al.(2024)Xu, Yao, Shu, Sun, Wu, Yu, Goldstein, and
  Huang]{xu2024shadowcast}
Yuancheng Xu, Jiarui Yao, Manli Shu, Yanchao Sun, Zichu Wu, Ning Yu, Tom
  Goldstein, and Furong Huang.
\newblock Shadowcast: Stealthy data poisoning attacks against vision-language
  models.
\newblock \emph{Advances in Neural Information Processing Systems},
  37:\penalty0 57733--57764, 2024.

\bibitem[Yang et~al.(2025)Yang, Li, Yang, Zhang, Hui, Zheng, Yu, Gao, Huang,
  Lv, et~al.]{yang2025qwen3}
An~Yang, Anfeng Li, Baosong Yang, Beichen Zhang, Binyuan Hui, Bo~Zheng, Bowen
  Yu, Chang Gao, Chengen Huang, Chenxu Lv, et~al.
\newblock Qwen3 technical report.
\newblock \emph{arXiv preprint arXiv:2505.09388}, 2025.

\bibitem[Zhang et~al.(2025)Zhang, Ye, Ma, Li, Yang, Chen, Sang, and
  Yeung]{zhang2025anyattack}
Jiaming Zhang, Junhong Ye, Xingjun Ma, Yige Li, Yunfan Yang, Yunhao Chen, Jitao
  Sang, and Dit-Yan Yeung.
\newblock Anyattack: Towards large-scale self-supervised adversarial attacks on
  vision-language models.
\newblock In \emph{Proceedings of the Computer Vision and Pattern Recognition
  Conference}, pages 19900--19909, 2025.

\bibitem[Zhang et~al.(2018)Zhang, Isola, Efros, Shechtman, and
  Wang]{zhang2018unreasonable}
Richard Zhang, Phillip Isola, Alexei~A Efros, Eli Shechtman, and Oliver Wang.
\newblock The unreasonable effectiveness of deep features as a perceptual
  metric.
\newblock In \emph{Proceedings of the IEEE conference on computer vision and
  pattern recognition}, pages 586--595, 2018.

\bibitem[Zhao et~al.(2025)Zhao, Duan, Wang, Chen, Kang, Ruan, Tao, Chen, Xue,
  and Wei]{zhao2025jailbreaking}
Shiji Zhao, Ranjie Duan, Fengxiang Wang, Chi Chen, Caixin Kang, Shouwei Ruan,
  Jialing Tao, YueFeng Chen, Hui Xue, and Xingxing Wei.
\newblock Jailbreaking multimodal large language models via shuffle
  inconsistency.
\newblock In \emph{Proceedings of the IEEE/CVF International Conference on
  Computer Vision}, pages 2045--2054, 2025.

\bibitem[Zhao et~al.(2023)Zhao, Pang, Du, Yang, Li, Cheung, and
  Lin]{zhao2023evaluating}
Yunqing Zhao, Tianyu Pang, Chao Du, Xiao Yang, Chongxuan Li, Ngai-Man~Man
  Cheung, and Min Lin.
\newblock On evaluating adversarial robustness of large vision-language models.
\newblock \emph{Advances in Neural Information Processing Systems},
  36:\penalty0 54111--54138, 2023.

\bibitem[Zheng et~al.(2023)Zheng, Chiang, Sheng, Zhuang, Wu, Zhuang, Lin, Li,
  Li, Xing, et~al.]{zheng2023judging}
Lianmin Zheng, Wei-Lin Chiang, Ying Sheng, Siyuan Zhuang, Zhanghao Wu, Yonghao
  Zhuang, Zi~Lin, Zhuohan Li, Dacheng Li, Eric Xing, et~al.
\newblock Judging llm-as-a-judge with mt-bench and chatbot arena.
\newblock \emph{Advances in neural information processing systems},
  36:\penalty0 46595--46623, 2023.

\bibitem[Zhou et~al.(2025)Zhou, Wang, Li, Lin, Liu, Dong, and Guo]{dps}
Qi~Zhou, Dongxia Wang, Tianlin Li, Yun Lin, Yang Liu, Jin~Song Dong, and Qing
  Guo.
\newblock Defending {LVLM}s against vision attacks through partial-perception
  supervision.
\newblock In \emph{Forty-second International Conference on Machine Learning},
  2025.
\newblock URL \url{https://openreview.net/forum?id=C4F42Ho7IM}.

\end{thebibliography}
\end{document}